\documentclass[11pt]{article}
\usepackage{appendix}
\usepackage{amsmath,amssymb,amscd,amsthm}
\usepackage{graphicx}
\usepackage{dcolumn}
\usepackage{bm}
\usepackage{ifthen}
\usepackage{rays_defs_18}

\usepackage{hyperref}
\hypersetup{
    bookmarks=true,         
    unicode=false,          
    pdftoolbar=true,        
    pdfmenubar=true,        
    pdffitwindow=false,     
    pdfstartview={FitH},    
    pdfauthor={Reinhard Heckel},     
    pdfsubject={Subject},   
    pdfcreator={Reinhard Heckel},   
    pdfproducer={Producer}, 
    pdfnewwindow=true,      
    colorlinks=true,       
    linkcolor=red,          
    citecolor=green,        
    filecolor=magenta,      
    urlcolor=cyan           
}

\usepackage{filecontents}

\usepackage[
backend=bibtex,
url=false,isbn=false,doi=false,
date=year,
hyperref=auto,
style=alphabetic,
natbib=true,
sorting=nty,
firstinits=true,
maxnames=2, maxbibnames=10,
]{biblatex}

\bibliography{./bibliography.bib} 

\usepackage{bbm}
\newcommand\ind[1]{ \mathbbm{1}_{\left\{ #1 \right\}} }

\usepackage{comment}

\usepackage{tikz}
\usepackage{pgfplots}
\usepgfplotslibrary{groupplots} 
\usetikzlibrary{pgfplots.groupplots}
\usetikzlibrary{matrix,arrows,decorations.pathmorphing}
\usepackage{pgfplots,pgfplotstable}
\usepgfplotslibrary{fillbetween}
\pgfplotsset{compat=1.10}

\usetikzlibrary{matrix,arrows,decorations.pathmorphing}

\usepackage{amsmath,amssymb,ifthen,color,framed,bm}

\usepackage{colortbl}
\definecolor{tblblue}{RGB}{101,124,191}
\definecolor{tblred}{rgb}{1,0.93,0.93}
\definecolor{DarkBlue}{rgb}{0,0,0.7} 
\definecolor{BrickRed}{RGB}{203,65,84}

\newtheorem{lemma}{Lemma}

\newtheorem{theorem}{Theorem}

\newtheorem{proposition}{Proposition}

\newcommand\loss{\ell}
\newcommand\stepsize{\eta}
\newcommand\iter{t}
\newcommand\iteralt{\tau} 

\newcommand\risk{R}
\newcommand\emprisk{\hat R}

\newcommand\ball{\mc B}

\newcommand\vtheta{{\bm \theta}}

\newcommand\vepsilon{{\bm \epsilon}}

\newcommand\mSigma{\bm \Sigma}
\newcommand\mEta{\diag(\bm \eta)}

\newcommand\setD{\mathcal D}
\newcommand\setF{\mathcal F}
\newcommand\setR{\mathcal R}
\newcommand\setW{\mathcal W}
\newcommand\setV{\mathcal V}

\newcommand\Jac{\mc J}
\newcommand\mystackrel[2]{\stackrel{\text{#1}}{#2}}

\newcommand\relu{\mathrm{relu}}

\long\def\comment#1{}

\newcommand\vectorize[1]{\mathrm{vect}(#1)}

\pgfplotsset{select coords between index/.style 2 args={
    x filter/.code={
        \ifnum\coordindex<#1\fi
        \ifnum\coordindex>#2\fi
    }
}}

\begin{document}

\begin{center}

{\bf{\LARGE{
Early Stopping in Deep Networks: Double Descent and How to Eliminate it
}}}

\vspace*{.2in}

{\large{
\begin{tabular}{cccc}
Reinhard Heckel$^{\dagger,\ast}$ and Fatih Furkan Yilmaz$^{\ast}$
\end{tabular}
}}

\vspace*{.05in}

\begin{tabular}{c}
$^\dagger$Dept. of Electrical and Computer Engineering, Technical University of Munich\\
$^\ast$Dept. of Electrical and Computer Engineering, Rice University
\end{tabular}

\vspace*{.1in}

\today

\vspace*{.1in}

\end{center}


\begin{abstract}
Over-parameterized models, such as large deep networks, often exhibit a double descent phenomenon, where as a function of model size, error first decreases, increases, and decreases at last. This intriguing double descent behavior also occurs as a function of training epochs, and has been conjectured to arise because training epochs control the model complexity. In this paper, we show that such epoch-wise double descent arises for a different reason: It is caused by a superposition of two or more bias-variance tradeoffs that arise because different parts of the network are learned at different epochs, and eliminating this by proper scaling of stepsizes can significantly improve the early stopping performance.
We show this analytically for
i) linear regression, where differently scaled features give rise to a superposition of bias-variance tradeoffs, and for
ii) a two-layer neural network, where the first and second layer each govern a bias-variance tradeoff. 
Inspired by this theory, we study two standard
convolutional networks empirically, and show that eliminating epoch-wise double descent through adjusting stepsizes of different layers improves the early stopping performance significantly. 
\end{abstract}


\section{Introduction}

Most machine learning algorithms learn a function that predicts a label from features. 
This function lies in a hypothesis class, such as a neural networks 
parameterized by its weights.
Learning amounts to fitting the parameters of the function by minimizing an empirical risk over the training examples. 
The goal is to learn a function that performs well on new examples, which are assumed to come from the same distribution as the training examples.

Classical machine learning theory says that the test error or risk as a function of the size of the hypothesis class is U-shaped: a small hypothesis class is not sufficiently expressive to have small error, and a large one leads to overfitting to spurious patterns in the data. 
The superposition of those two sources of errors, typically referred to as bias and variance, yields the classical U-shaped curve.

However, increasing the model size beyond the number of training examples decreases the error again. 
This phenomena, dubbed ``double descent'' by \citet{belkin_ReconcilingModernMachinelearning_2019} has been observed as early as 1995 by~\citet{opper_StatisticalMechanicsLearning_1995}, and is highly relevant today because most modern machine learning models, in particular deep neural networks, operate in the over-parameterized regime, where the error decreases again as a function of model size, and where the 
model is sufficiently expressive to describe any data, even noise.

\begin{figure}

\begin{center}
\begin{tikzpicture}[>=latex]

\draw[red,dashed,<-] (2.3cm,0cm) -- (2.3cm,3.6cm) node [above] {\small best early stopping time};

\draw[red,dashed,<-] (12.1cm,0cm) -- (12.1cm,3.6cm) node [above] {\small best early stopping time};

\node at (2.3cm,1.6cm) [red,circle,fill,inner sep=1.5pt]{};
\node at (12.1cm,1.35cm) [red,circle,fill,inner sep=1.5pt]{};

\begin{groupplot}[
width=8cm,
height=5cm,
group style={
    group name=multi_dd,
    group size= 2 by 1,
    xlabels at=edge bottom,
    ylabels at=edge left,
    xticklabels at=edge bottom,
    horizontal sep=1.8cm, 
    vertical sep=1.8cm,
    },
legend cell align={left},
legend columns=1,
legend style={
            at={(1,1)}, 
            anchor=north east,
            draw=black!50,
            fill=none,
            font=\small
        },
colormap name=viridis,
cycle list={[colors of colormap={50,350,750}]},
tick align=outside,
tick pos=left,
x grid style={white!69.01960784313725!black},
xtick style={color=black},
y grid style={white!69.01960784313725!black},
ytick style={color=black},
xlabel={epoch}, 
xmode=log,
]


\nextgroupplot[ymin=0, ymax=0.42, ylabel={train/test error}]

\addplot +[mark=none] table[x=epoch,y expr=1 - \thisrow{test}/100]{fig/data/double_descent_5layer.txt};\addlegendentry{test};
\addplot +[mark=none] table[x=epoch,y expr=1 - \thisrow{train}/100]{fig/data/double_descent_5layer.txt};
\addlegendentry{train};

\nextgroupplot[xlabel = {$\iter$ iterations},ylabel = {risk},xmin=1]
      \addplot + [thick,mark=none,dash pattern=on 1pt off 2pt on 3pt off 2pt] table[x index=0,y index=1]{./fig/risk_separate.dat};   
      \addlegendentry{bias-variance 1};
      \addplot + [thick,mark=none,dash pattern=on 1pt off 1pt on 1pt off 1pt] table[x index=0,y index=2]{./fig/risk_separate.dat};
      \addlegendentry{bias-variance 2};
      \addplot + [thick,mark=none] table[x index=0,y index=4]{./fig/risk_separate.dat};
      \addlegendentry{risk};

\end{groupplot}

\end{tikzpicture}

\end{center}

\vspace{-0.6cm}

\caption{
\label{fig:doubledescentexample}
{\bf Left:} The test error of an over-parameterized 5-layer convolutional network trained on a noisy version of the CIFAR-10 training set (20\% random label noise) as a function of training epochs. As observed by~\citet{nakkiran_DeepDoubleDescent_2020}, the performance shows a double descent behavior.
{\bf Right:}
As we show in this paper, the risk of a regression problem can be decomposed as the sum of two bias-variance tradeoffs.
{\bf Both examples:} Early stopping the training where the test error achieves its minima is critical for performance.
}
\end{figure}
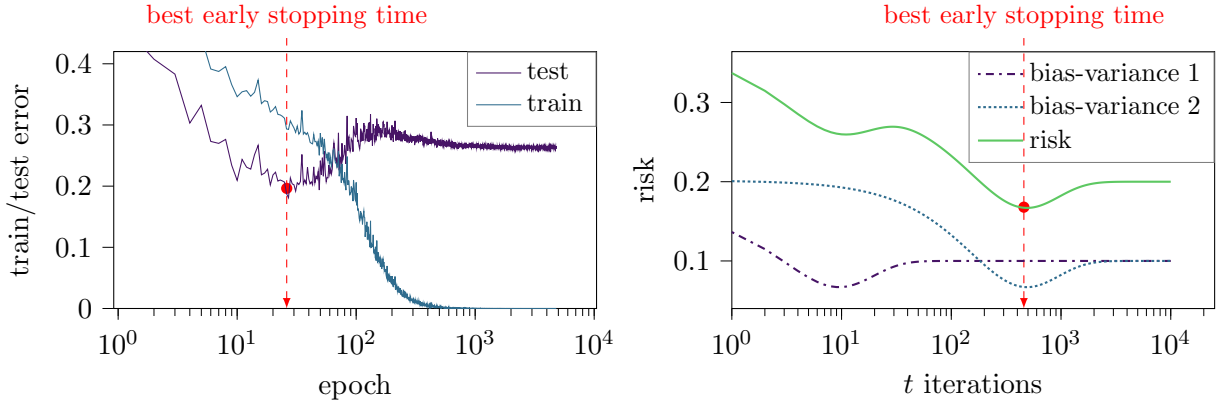

Interestingly, this double descent behavior also occurs as a function of training time, as observed by~\citet{nakkiran_DeepDoubleDescent_2020} and as illustrated in Figure~\ref{fig:doubledescentexample}, in particular for learning from noisy labels. 
The left panel of Figure~\ref{fig:doubledescentexample} shows that as a function of training epochs,
the test error first decreases, increases, and then decreases again. It is critical to understand this so-called \emph{epoch-wise double descent} behavior to determine the early stopping time that gives the best performance. Early stopping, or other regularization techniques, are critical for learning from noisy labels~\cite{arpit_CloserLookMemorization_2017a,yilmaz_ImageRecognitionRaw_2020}. 

\citet{nakkiran_DeepDoubleDescent_2020} conjectured that epoch-wise double descent occurs because the training time controls the ``effective model complexity''. 
This conjecture is intuitive, because the model-size, and thus the size of the hypothesis class, can be controlled 
by regularizing the empirical risk via early stopping the gradient descent iterations, as formalized in the under-parameterized regime by~\citet{yao_EarlyStoppingGradient_2007a,raskutti_EarlyStoppingNonparametric_2014,buhlmann_BoostingL2Loss_2003}. Specifically, limiting the number of gradient descent iterations ensures that the functions parameters lie in a ball around the initial parameters, and thereby limits the size of the hypothesis class.

In this paper, however, we show empirically and theoretically that epoch-wise double descent 
arises for a different reason: It is caused by a superposition of bias-variance tradeoffs, as illustrated for a toy-regression example in the right panel of Figure~\ref{fig:doubledescentexample}. 
If the risk can be decomposed into two U-shaped bias-variance tradeoffs with minima at different epochs/iterations, then the overall risk/test error has a double descent behavior.

Contrary to model-wise double descent, epoch-wise double descent is not a phenomena tied to over-parameterization. Both under- and overparamterized models can have epoch-wise double descent as we show in this paper.


\subsection{Contributions}

The goal of this paper is to understand the epoch-wise double descent behavior observed by~\citet{nakkiran_DeepDoubleDescent_2020} and depicted in Figure~\ref{fig:doubledescentexample}. 
Our main finding is that contrary to model-wise double descent, epoch-wise double descent is not a result of controlling the model complexity with early stopping, but a consequence of a superposition of bias variance tradeoffs. Our contributions are as follows:
\begin{itemize}
\item 
First, we consider a linear regression model and theoretically characterize the risk of early stopped least squares. We show that if features have different scales, then the early stopped least squares estimate as a function of the early stopping time is a superposition of bias-variance tradeoffs, which yields a double descent like curve. See Figure~\ref{fig:doubledescentexample}, right panel, as an illustration.
\item 
Second, we characterize the early stopped risk of a two-layer neural network theoretically and show that it is upper bounded by a curve consisting of over-lapping bias-variance tradeoffs that are governed by the initializations and stepsizes of the two layers. Specifically, the initialization scales and stepsizes of the weights in the first and second layer determine whether double descent occurs or not.
We provide numerical examples showing how epoch-wise double descent occurs when training such a two-layer network on data drawn from a linear model, and how it can be eliminated by scaling the stepsizes of the layers accordingly.
\item 
Third, we study a standard 5-layer convolutional network as well as ResNet-18 
empirically. For the 5-layer convolutional network we find---similarly as for the two-layer model---epoch-wise double descent occurs because the convolutional layers (representation layers) are learned faster than the final, fully connected layer, which results in a superposition of bias-variance tradeoffs. 
Similarly, for ResNet-18, we find that later layers are learned faster than early layers, which again results in double descent. 
In both cases, epoch-wise double descent can be eliminated through adjusting the stepsizes of different coefficients or layers.
%
\end{itemize}

In summary, we provide new examples on when epoch-wise double descent occurs, as well as analytical results explaining epoch-wise double descent theoretically. Our theory is constructive in that it suggests a simple and effective mitigation strategy: scaling stepsizes appropriately. 
We also note that epoch-wise double descent should be eliminated by adjusting the stepsizes and/or the initialization, because this often translates to better overall performance.




\subsection{Related works}

There is a large number of works that have studied early stopping theoretically. 
Intuitively, each step of an iterative algorithm reduces the bias but increases variance. Thus early stopping can ensure that neither bias nor variance are too large. A variety of papers~\cite{yao_EarlyStoppingGradient_2007a,raskutti_EarlyStoppingNonparametric_2014,buhlmann_BoostingL2Loss_2003,wei_EarlyStoppingKernel_2019} formalized this intuition and developed theoretically sound early stopping rules. 
Those works do not, however, predict that a double descent curve can occur.

A second, more recent line of works, have studied early stopping from a different perspective, namely that of gradient descent fitting different components of a signal or different labels at different speeds. For a linear least squares problem, the data in the direction of singular vectors associated with large singular values is fitted faster than that in the direction of singular vectors associated with small singular values.
\cite{advani_HighdimensionalDynamicsGeneralization_2017a} have shown this for a linear least squares problem or stated differently, a linear neural network with a single layer.
\citet{li_GradientDescentEarly_2020,arora_FineGrainedAnalysisOptimization_2019} have shown that this view explains why neural network often fit clean labels before noisy ones, and \citet{heckel_DenoisingRegularizationExploiting_2020} have used this view to prove that convolutional neural networks provably denoise images. 

Our theoretical results for neural networks build on a line of works that relate the dynamics of gradient descent to those of an associated linear model or a kernel method in the highly overparameterized regime~\cite{jacot_NeuralTangentKernel_2018,lee_DeepNeuralNetworks_2018a,arora_FineGrainedAnalysisOptimization_2019,du_GradientDescentProvably_2018b,
oymak_ModerateOverparameterizationGlobal_2020,
oymak_GeneralizationGuaranteesNeural_2019,heckel_DenoisingRegularizationExploiting_2020}.
We use the same proof strategy as those papers to characterize the early stopping performance of a simple two-layer neural network, but in contrast to those earlier works, we develop early stopping results and optimize over both the weights in the first and second layer, as opposed to only optimizing over the weights in the first layer. That is important, because we want to demonstrate that initialization and stepsize choices of different layers lead to different bias-variance tradeoffs.

Next, we note that there is an emerging line of works that theoretically establishes double descent behavior as a function of the model complexity (e.g., measured by the number of parameters of the model) for linear regression~\cite{hastie_SurprisesHighDimensionalRidgeless_2019,belkin_TwoModelsDouble_2019}, for random feature regression~\cite{mei_GeneralizationErrorRandom_2019,dascoli_DoubleTroubleDouble_2020}, and for binary linear regression~\cite{deng_ModelDoubleDescent_2020}. 

A number of recent theoretical double-descent works~\cite{jacot_ImplicitRegularizationRandom_2020,yang_RethinkingBiasVarianceTradeoff_2020,dascoli_DoubleTroubleDouble_2020} have decomposed the risk into bias and variance terms, and studied their behavior. Those works demonstrate that the bias typically decreases as a function of the model size, and the variance first increases, and then decreases, which can yield to a double-descent behavior. 
As we demonstrate in Appendix~\ref{sec:bias-variancedecomposition}, the epoch-wise  double descent phenomena for the standard CNN studied by~\citet{nakkiran_DeepDoubleDescent_2020} \emph{cannot be explained} with a similar behavior of bias and variance as a function of the training time: The variance is increasing as a function of training epochs, as opposed to being unimodal. 
Also, contrary to epoch-wise double descent studied in this paper, model wise double descent is \emph{not explained} in the literature by a superposition of bias variance tradeoffs.

Finally, our suggestion to mitigate epoch-wise double descent with step-size adaption and early stopping is a form of regularization. 
Related work for model-wise double descent shows that model-wise double descent can be mitigated with ($\ell_2$) regularization~\cite{nakkiran_OptimalRegularizationCan_2020}. 

\section{Early-stopped gradient descent for linear least squares}
\label{sec:leastsquares}

We start by studying the risk of early stopped gradient descent for fitting a linear model to data generated by a Gaussian linear model. Our main finding is that the risk as a function of the early stopping time is characterized by a superposition of U-shaped bias-variance tradeoffs, and if the features of the Gaussian linear model have different scales, those bias-variance tradeoff curves add up to a double descent shaped risk curve. 
We also show that the early stopping performance of the estimator can be improved through double descent elimination by 
scaling the stepsizes associated with the features.

\subsection{Data model and risk}
\label{sec:datamodel}
Consider a regression problem, and suppose data is generated from a Gaussian linear model as
\[
y = \innerprod{\vx}{\vtheta^\ast} + z,
\]
where $\vx \in \reals^d$ is a zero-mean Gaussian feature vector with diagonal co-variance matrix $\mSigma = \diag(\sigma_1^2,\ldots,\sigma_d^2)$, and $z$ is independent, zero-mean Gaussian noise with variance $\sigma^2$.
We are given a training set $\setD = \{(\vx_1,y_1),\ldots,(\vx_n,y_n)\}$ consisting of $n$ data points drawn iid from this Gaussian linear model. 

We consider the class of linear estimators parameterized by a vector $\hat \vtheta \in \reals^d$, which we estimate based on the training data $\setD$. The linear estimator predicts the label associated with a feature vector $\vx$ as $\hat y = \innerprod{\vx}{\hat \vtheta}$.
The (mean-squared) risk of this estimator is
\begin{align*}
\risk(\hat \vtheta)
&=
\EX{\left(y - \innerprod{\vx}{\hat \vtheta}\right)^2},
\end{align*}
where expectation is over an example $(\vx,y)$ drawn independently (of the training set) from the underlying linear model. 
The risk of the estimator can be written as a function of the variances of the features, $\sigma_i^2$, and of the coefficients of the underlying true linear model, $\vtheta^\ast = [\theta_1^\ast,\ldots, \theta_d^\ast]$, as
\begin{align}
\risk(\hat \vtheta)
%
%
&=
\sigma^2
+
\sum_{i=1}^d \sigma_i^2 (\theta^\ast_i - \hat \theta_i)^2.
\label{eq:expressioniterates}
\end{align}

\subsection{Early-stopped least squares estimate}

We consider the estimate based on early stopping gradient descent applied to the empirical risk
\[
\emprisk(\vtheta) = \norm[2]{\mX \vtheta - \vy}^2.
\]
Here, the matrix $\mX \in \reals^{n\times d}$ contains the scaled training feature vectors $\frac{1}{\sqrt{n}}\vx_1,\ldots, \frac{1}{\sqrt{n}} \vx_n$ as rows, and $\vy = \frac{1}{\sqrt{n}}[y_1,\ldots,y_n]$ are the corresponding scaled responses. 
We initialize gradient descent with $\vtheta^0 = \bm 0$ and iterate, for $\iter=1,2,\ldots$,
\[
\vtheta^{\iter+1} 
= \vtheta^{\iter} -   \frac{1}{2} \mEta \nabla \emprisk(\vtheta^{\iter}),
\]
where $\mEta$ is a diagonal matrix containing the stepsizes $\eta_i>0$ associated with each of the features as entries. Note that we allow for different stepsizes for all of the features. 
In the following, we study the properties of the estimate obtained by early stopping gradient descent at iteration $\iter$, i.e., $\vtheta^{\iter}$.

\subsection{Risk of early stopped least squares}
\label{sec:leastsquares}

The main result of this section is that in the underparameterized regime, where $d \ll n$, the risk of gradient descent after $\iter$ iterations, $\risk(\vtheta^{\iter})$, is very close to a risk expression defined as
\begin{align}
\label{eq:defbarR}
\bar R(\tilde \vtheta^{\iter})
\defeq
\sigma^2 +
\sum_{i=1}^d 
\underbrace{
\sigma_i^2
(\theta^\ast_i)^2
(1 - \eta_i \sigma_i^2)^{2\iter}  
+  
\frac{\sigma^2}{n} (1 - (1-\eta_i \sigma_i^2)^\iter)^2
}_{U_i(\iter)},
\end{align}
as formalized be the theorem below. 

\begin{theorem}
\label{thm:maindifference}
Suppose that the stepsizes obey $\eta_i \leq \frac{1}{\sigma_i^2}$, for all $i=1,\ldots,d$. 
With probability at least
$1 - 2d^{-5} + 2de^{-n/8} - e^{-d} - 2e^{-32}$
over the random training set generated by a linear Gaussian model with parameters $\vtheta^\ast$ and  $\mSigma$,
the difference of the early stopped risk and the risk expression in~\eqref{eq:defbarR} is at most 
\begin{align}
\left| \risk(\vtheta^{\iter}) - \bar R(\tilde \vtheta^{\iter}) \right|
\leq
c
\left(
\frac{\max_i \eta_i^2 \sigma_{i}^4}{\min_i \eta_i \sigma_{i}^4} 
\frac{d}{n}
\left(
\norm[2]{\mSigma \vtheta^{\ast} }^2 
+
\frac{d}{n} \sigma^2 \log(d)
\right)
+
\frac{\sigma^2}{n} \sqrt{d}
\right).
\end{align}
Here, $c$ is a numerical constant.
\end{theorem}

Theorem~\ref{thm:maindifference} guarantees that with high probability the risk $\risk(\vtheta^{\iter})$ is well approximated by the risk expression $\bar R(\tilde \vtheta^{\iter})$, provided the model is sufficiently underparameterized (i.e., $d/n$ is large).

\begin{figure}

\begin{center}
\begin{tikzpicture}[>=latex]

\draw[red,dashed,<-] (3.0cm,0cm) -- (3.0cm,3.4cm) node [above] {\small best stopping time};

\draw[red,dashed,<-] (6.3cm,0cm) -- (6.3cm,3.4cm) node [above] {\small best stopping time};

\draw[red,dotted] (3.0cm,1.1cm) -- (6.3cm,1.1cm);

\node at (3.0cm,1.35cm) [red,circle,fill,inner sep=1.5pt]{};
\node at (6.3cm,1.1cm) [red,circle,fill,inner sep=1.5pt]{};

\begin{groupplot}[
width=6.0cm,
height=4.8cm,
group style={
    group name=multi_dd,
    group size= 3 by 1,
    xlabels at=edge bottom,
    ylabels at=edge left,
    yticklabels at=edge left,
    xticklabels at=edge bottom,
    horizontal sep=0.8cm, 
    vertical sep=0.8cm,
    },
legend cell align={left},
legend columns=1,
legend style={
            at={(1,1)}, 
            anchor=north east,
            draw=black!50,
            fill=none,
            font=\small
        },
colormap name=viridis,
cycle list={[colors of colormap={50,350,750}]},
tick align=outside,
tick pos=left,
x grid style={white!69.01960784313725!black},
xtick style={color=black},
y grid style={white!69.01960784313725!black},
ylabel style={at={(-0.20,0.5)}},
ytick style={color=black},
xlabel={epoch}, 
xmode=log,
ymax=0.4,
ymin=0,
xmax = 10000,
title style={at={(0.5,-0.65)}},
]


\nextgroupplot[xlabel = {$\iter$ iterations},ylabel = {risk},xmin=1,title={\small a) constant stepsize}
]
      \addplot + [thick,mark=none,dash pattern=on 1pt off 2pt on 3pt off 2pt] table[x index=0,y index=1]{./fig/risk_separate.dat};   
      \addlegendentry{bias-variance 1};
      \addplot + [thick,mark=none,dash pattern=on 1pt off 1pt on 1pt off 1pt] table[x index=0,y index=2]{./fig/risk_separate.dat};
      \addlegendentry{bias-variance 2};
      \addplot + [thick,mark=none] table[x index=0,y index=4]{./fig/risk_separate.dat};
      \addlegendentry{1+2};


\nextgroupplot[xlabel = {$\iter$ iterations},xmin=1,title={\small b) elimination with diff. stepsizes}
]
      \addplot + [thick,mark=none,dash pattern=on 1pt off 2pt on 3pt off 2pt] table[x index=0,y index=1]{./fig/risk_separate.dat};   
      \addlegendentry{bias-variance 1};
      \addplot + [thick,mark=none,dash pattern=on 2pt off 3pt on 2pt off 1pt] table[x index=0,y index=3]{./fig/risk_separate.dat};
      \addlegendentry{bias-variance 3};
      \addplot + [thick,mark=none] table[x index=0,y index=5]{./fig/risk_separate.dat};
      \addlegendentry{1+3};
      %


\nextgroupplot[xlabel = {$\iter$ iterations},xmin=1,title={\small c) before \& after elimination}
]

      \addplot + [thick,mark=none,green!70!black,dashed] table[x index=0,y index=4]{./fig/risk_separate.dat};
      \addlegendentry{1+2};
      
      \addplot + [thick,mark=none,green!70!black] table[x index=0,y index=5]{./fig/risk_separate.dat};
      \addlegendentry{1+3};
      %

\end{groupplot}

\end{tikzpicture}
\end{center}

\vspace{-0.4cm}

\caption{
Early stopped least squares risk for a two-feature Gaussian linear model.
\label{fig:mitigatingdouble}
{\bf a:}
Two U-shaped bias-variance tradeoffs $U_{i}(\iter)$ defined in~\eqref{eq:defbarR} for the parameters 
$\theta_1^\ast = 1.5, \sigma_1 = 1, \eta_1 = 0.05$ (bias-variance 1) and for the parameters 
$\theta_2^\ast = 10, \sigma_2 = 0.15, \eta_2 = 0.05$
(bias-variance 2), along with their sum (1+2) which determines the risk. 
{\bf b:}
Same plot, but this time the bias-variance tradeoff $U_{2}(\iter)$ is shifted to the left by increasing the stepsize $\eta_2$ according to Proposition~\ref{prop:mincond} (yielding bias-variance tradeoff 3), so that its minimum overlaps with that of bias-variance tradeoff 1.
This eliminates the double descent behavior and gives a better performance of the resulting estimator (a smaller risk or error).
{\bf c:} The resulting risk curves before and after elimination, demonstrating that the minimum of the risk after double descent elimination is smaller than before elimination.
}
\end{figure}
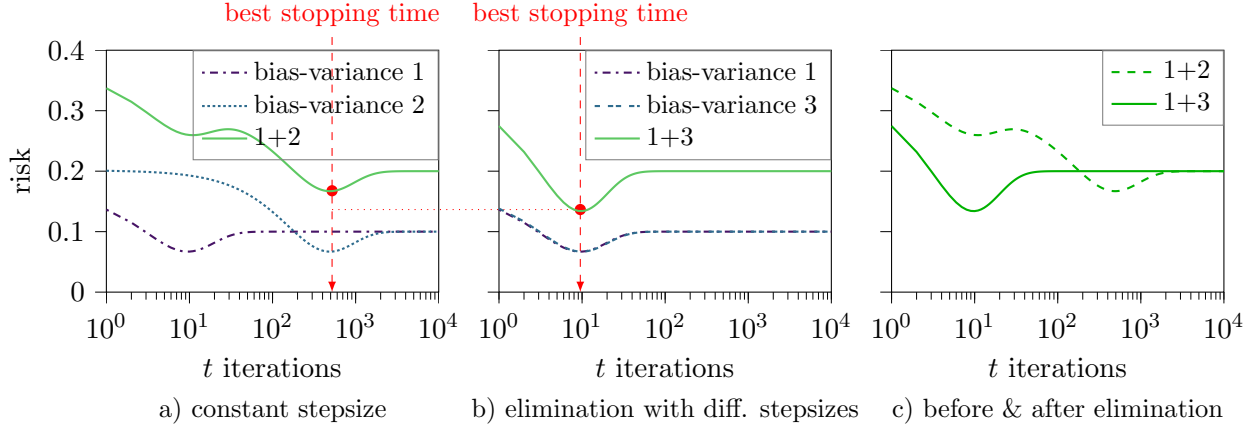

As a consequence, the risk of early stopped least-squares is a superposition of U-shaped bias-variance tradeoffs, and if the features are differently scaled, this can give rise to epoch-wise double descent. 
To see this, first note that the terms $U_i(\iter)$ in the risk expression~\eqref{eq:defbarR} are U-shaped as a function of the early stopping time $\iter$, because
$\sigma_i^2(\theta^\ast_i)^2
(1 - \eta_i \sigma_i^2)^{2\iter}$ decreases in $\iter$ and $\frac{\sigma^2}{n} (1 - (1-\eta_i \sigma_i^2)^\iter)^2$ increases in $\iter$; see Figure~\ref{fig:mitigatingdouble}a for an example.
The minima of the individual U-shaped curves $U_i(\iter)$ depend on the product of the stepsize and the $i$-th features' variance, $\eta_i \sigma_i^2$; the larger this product, the earlier (as a function of the number of iterations, $\iter$) the respective U-shaped curve reaches its minimum. 
Therefore, if we add up two (or more) such U-shaped curves with minima at different iterations, the resulting risk curve can have a double descent shape (again, see Figure~\ref{fig:mitigatingdouble}a for an example). 
This establishes our claim that differently scaled features in linear regression give rise to epoch-wise double descent. 

\paragraph{Improving performance by eliminating double descent:}
Epoch-wise double descent can be eliminated or mitigated by properly scaling the stepsizes associated with each of the features, so that the individual bias-variance tradeoffs overlap at the same point, as formalized by the following proposition:

\begin{proposition}
\label{prop:mincond}
Pick an optimal early stopping time $\tilde \iter \geq 1$. 
The minimum of the risk expression 
$\min_{\eta_1,\ldots,\eta_d} \min_\iter \bar R(\tilde \vtheta^{\iter})$ is achieved at iteration $\tilde \iter$ by choosing the stepsizes pertaining to the features as
\[
\eta_i = \frac{1}{\sigma_i^2} \left( 1 - \left( \frac{\sigma^2 /n }{ \sigma_i^2 (\theta_i^\ast)^2 + \sigma^2 /n }  \right)^{1/\tilde \iter} \right).
\]
\end{proposition}

Proposition~\ref{prop:mincond} is proven by showing that if the stepsizes are chosen as specified as above, then the minima of the U-shaped curves $U_i(\iter)$ occur at the same iteration $\tilde \iter$. 

Elimination of double descent is illustrated in Figure~\ref{fig:mitigatingdouble}b, where we choose the stepsizes so that the two U-shaped curves overlap. 
By eliminating double descent optimally so that all the individual bias-variance tradeoffs $U_{i}(\iter)$ achieve their minima at the same early stopping point $\iter$, we  achieve the lowest overall risk at the optimal early stopping point. Therefore eliminating double descent is critical for optimal performance. 

In practice we may not know the variances of the features exactly, and therefore may not be able to optimally choose the stepsizes. However, we may be able to mitigate double descent sub-optimally by treating the stepsizes as hyperparameters and scaling them appropriately.


\paragraph{Intuition for the risk expression~\eqref{eq:defbarR}:}
The appendix contains a proof of Theorem~\ref{thm:maindifference}. Here we provide intuition why the risk is governed by the risk expression~\eqref{eq:defbarR}. 
The gradient descent iterates obey
\begin{align*}
\vtheta^{\iter+1} - \vtheta^\ast 
=
\left( \mI - \mEta \transp{\mX}\mX \right) (\vtheta^{\iter} - \vtheta^\ast)
+ \mEta \transp{\mX}\vz,
\end{align*}
where $\vz = [z_1,\ldots, z_n]$ is the noise. 
As we formalize below, in the under-parameterized regime where $n \gg d$, we have that $\transp{\mX} \mX \approx \mSigma^2$. Therefore the original iterates are close to the proximal iterates $\tilde \vtheta^{\iter}$ defined by
\begin{align}
\label{eq:closeiterates}
\tilde \vtheta^{\iter+1} - \vtheta^\ast 
=
\left( \mI - \mEta \transp{\mSigma} \mSigma \right) (\tilde \vtheta^{\iter} - \vtheta^\ast)
+
\mEta \transp{\mX}\vz.
\end{align}
The proximal iterates are, up to the extra term $\mEta \transp{\mX}\vz$, equal to the iterates of gradient descent applied to the population risk $\risk(\vtheta)$. 
Note that in contrast to the literature where it is common to bound the deviation of the original iterates from the iterates on the population risk~\cite{raskutti_EarlyStoppingNonparametric_2014}, here we control the deviation of the original iterates to the proximal iterates $\tilde \vtheta^{\iter}$. 

The iterates $\tilde \vtheta^{\iter}$ can easily be written out in closed form. To do so, first note that for the recursion $\theta^{\iter+1} = \alpha \theta^{\iter} + \gamma$ we have
$
\tilde \theta^\iter
=
\alpha^\iter \theta^0 + \gamma \sum_{i=1}^{\iter-1}\alpha^i 
=
\alpha^\iter \theta^0 + \gamma \frac{1 - \alpha^\iter}{1-\alpha},
$
where we used the formula for a geometric series. 
Using this relation, and that we are starting our iterations at $\theta^0_i = 0$, we obtain for the $i$-th entry of $\tilde \vtheta^{\iter}$ that
\[
\tilde \theta^\iter_i - \theta^\ast_i 
=
(1 - \eta_i \sigma_i^2)^\iter \theta^\ast_i + \sigma_i \transp{\tilde \vx}_i \vz \frac{1 - (1-\eta_i \sigma_i^2)^\iter}{\sigma_i^2},
\]
where $\tilde \vx_i$ is the $i$-th \emph{column} of $\mX$ (not the $i$-th example/feature vector!). 
Next note that, $\EX{ (\transp{\tilde \vx}_i\vz)^2 } \approx \sigma^2 \sigma_i^2$ 
because the entries of $\vz$ are $\mc N(0,\sigma^2)$ distributed, and the entries of $\tilde \vx_i$ are $1/\sqrt{n} \mc N(0,\sigma_i^2)$ distributed. 
Using this expectation in the iterates $\tilde \vtheta^{\iter}$, and evaluating the risk of those iterates via the formula for the risk given by~\eqref{eq:expressioniterates} yields the risk expression~\eqref{eq:defbarR}. 
The proof of Theorem~\ref{thm:maindifference} in the appendix makes this intuition precise by formally bounding the difference of the proximal iterates to the original iterates.

\paragraph{Motivation for calling the U-shaped curves bias-variance tradeoffs:}
Let $\hat \vtheta = \hat \vtheta(\setD)$ be the parameter obtained based on the training data (for example by early stopping). 
The textbook bias-variance decomposition of the risk of $\hat \vtheta$ is
\begin{align*}
\EX[\setD]{\risk(\hat \vtheta)} 
=
\underbrace{ 
\EX[\vx]{ \left( \innerprod{\vx}{\vtheta^\ast} - \EX[\setD]{ \innerprod{\vx}{\hat \vtheta} } \right)^2 }
}_{\mathrm{Bias}(\hat \vtheta)}
+ 
\underbrace{ 
\EX[\setD,\vx]{ \left( \innerprod{\vx}{\hat \vtheta} - \EX[\setD]{\innerprod{\vx}{\hat \vtheta}} \right)^2 }
}_{\mathrm{Variance}(\hat \vtheta)}
+ \sigma^2. 
\end{align*}
The first term above 
is the bias of the hypothesis $\hat h(\vx) = \innerprod{\vx}{\hat \vtheta}$. It measures how well the average function can estimate the true underlying function $h(\vx) = \innerprod{\vx}{\vtheta^\ast}$. A low bias means that the hypothesis accurately estimates the true underlying function $\innerprod{\vx}{\vtheta^\ast}$. The second term is the variance of the method. The variance of the method measures the variance of the hypothesis over the training sets.

Recall from the previous paragraph that the estimate $\tilde \vtheta^t$ approximates the original iterations $\vtheta^t$ well provided that the model is sufficiently underparameterized, i.e., $d/n$ is small. 
It is straightforward to verify that
$\mathrm{Bias}(\tilde \vtheta^t)
=
\sum_{i=1}^d (\theta^\ast_i)^2
(1 - \eta_i \sigma_i^2)^{2\iter}  
$ 
and
$\mathrm{Variance}(\tilde \vtheta^t)
=
\sum_{i=1}^d \frac{\sigma^2}{n} (1 - (1-\eta_i \sigma_i^2)^\iter)^2
$, exactly equal to the bias and variance terms in the risk expression~\eqref{eq:defbarR}. 
It follows that the bias and variance of the original gradient descent iterates $\vtheta^t$ are also approximately equal to the terms in the risk expression~\eqref{eq:defbarR}. 
The U-shaped curves are then the bias-variance terms pertaining to the $i$-th feature; this formally establishes the U-shaped curves as bias-variance tradeoffs.

%

\section{Early stopping in two layer neural networks}



In this section, we establish a bound on the risk of a two-layer neural network and show that this bound can be interpreted as a super-position of U-shaped bias-variance tradeoffs, similar to the expression governing the risk of the linear model from the previous section. 
The risk of the two-layer network is governed by two associated kernels pertaining to the first and second layer, and the initialization scale and stepsizes of the weights in the first and second layer determine whether double descent occurs or not. We also show in an experiment that if double descent occurs, it can be eliminated by adapting the stepsizes of the two layers.

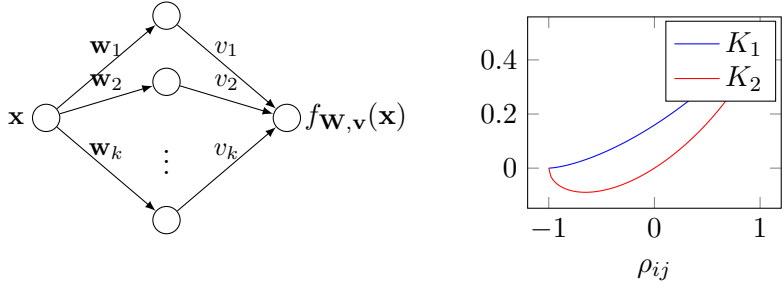
\begin{figure}[t]
\begin{center}
\begin{tikzpicture}

\begin{scope}[
scale=0.80,
xshift=-8.0cm,
yshift=1.5cm,
>=latex,
scale=1,
node/.style={circle,draw},
]

\node[node] (x) at (0,0) {};
\node at (-0.5,0) {\small $\vx$};
\node at (5.15,0) {$f_{\mW,\vv}(\vx)$};

\node[node] (a1) at (2,1.7) {};
\node[node] (a2) at (2,0.6) {};
\node[] 	   (a3) at (2,-0.6) {$\vdots$};
\node[node] (a4) at (2,-1.7) {};

\draw[->] (x) -- node[above] {\small $\vw_1$} (a1);
\draw[->] (x) -- node[above] {\small $\vw_2$} (a2);
\draw[->] (x) -- node[above] {\small $\vw_k$} (a4);

\node[node] (y) at (4,0) {};

\draw[->] (a1) -- node[above] {\small $v_1$} (y);
\draw[->] (a2) -- node[above] {\small $v_2$} (y);
\draw[->] (a4) -- node[above] {\small $v_k$} (y);

\end{scope}


\begin{groupplot}[
         title style={at={(0.5,-0.25)},anchor=north}, group
         style={group size=2 by 1, horizontal sep=1.2cm },
         width=0.3\textwidth,height=0.25\textwidth, xticklabel style={/pgf/number
           format/fixed, /pgf/number format/precision=3}, yticklabel
         style={/pgf/number format/fixed, /pgf/number
           format/precision=5}] 

\nextgroupplot[xlabel = {$\rho_{ij}$}
]
  \addplot + [mark=none] table[x index=0,y index=2]{./fig/NNkernels.dat};
  \addlegendentry{$K_1$};
       \addplot + [mark=none] table[x index=0,y index=1]{./fig/NNkernels.dat};
      \addlegendentry{$K_2$}; 
%
\end{groupplot}          
\end{tikzpicture}
\end{center}

\vspace{-0.75cm}
\caption{
{\bf Left:}
Two layer network model.
{\bf Right:} 
Kernel associated with the first, fully connected layer ($K_1$), and the output layer ($K_2$).
\label{fig:kernels}
}
\end{figure}

\paragraph{Network model:} We consider a two-layer neural network with ReLU activation functions and $k$ neurons in the hidden layer:
\begin{align}
\label{eq:networkmodel}
f_{\mW,\vv}(\vx) 
= \frac{1}{\sqrt{k}} \relu(\transp{\vx} \mW) \vv.
\end{align}
Here, $\vx \in \reals^d$ is the input of the network, $\mW \in \reals^{d \times k}$ are the weights of the first layer and $\vv \in \reals^k$ are the weights of the second layer. Moreover, $\relu(z) = \max(z,0)$ is the rectified linear unit, applied elementwise.
See the left panel of Figure~\ref{fig:kernels} for an illustration of this network.

\paragraph{Data model:} We assume that we are given a training set $\setD = \{(\vx_1,y_1), \ldots, (\vx_n,y_n)\}$ with examples $(\vx_i,y_i)$ drawn iid from some joint distribution. For convenience, we assume the datapoints are normalized, i.e., $\norm[2]{\vx_i} = 1$, and that the labels are bounded, i.e., $|y_i| \leq 1$. 

\paragraph{Training with early stopped gradient descent:}
We train the neural network with early stopped and randomly initialized gradient descent on a quadratic loss. We choose the weights at initialization $\vv^0,\mW^0$ as
\begin{align}
\label{eq:initialization}
[\mW^0]_{i,j} 
\sim \mc N(0, \omega^2), 
\quad 
[\vv^0]_i 
\sim \text{Uniform}(\{-\nu , \nu \}).
\end{align}
Here, $\omega$ and $\nu$ are parameters that trade off the magnitude of the weights of the first and second layer. 
Note that with this initialization, for a fixed unit norm feature vector $\vx$, we have $f_{\mW_0,\vv_0}(\vx) = O(\nu \omega)$. 
We apply gradient descent to the mean-squared loss
\[
\mc L(\mW,\vv)
=
\frac{1}{2} \sum_{i=1}^n(y_i - f_{\mW,\vv}(\vx_i))^2.
\]
The gradient descent updates are 
$\vv_{\iter+1} = \vv_{\iter} - \eta \nabla_\vv \mc L(\mW_\iter,\vv_\iter)$ and 
$\mW_{\iter+1} = \mW_{\iter} - \eta \nabla_\mW \mc L(\mW_\iter,\vv_\iter)$, where $\eta$ is a constant learning rate. 
We study the risk of the network as a function of the (early stopped) iterations $\iter$.

\paragraph{Evaluation and performance metric:}
Our goal is to bound the test error as a function of the iterations of gradient descent. Let $\loss \colon \reals \times \reals \to [0,1]$ be a loss function that is $1$-Lipschitz in its first argument and obeys $\loss(y,y) = 0$; a concrete example is the loss $\loss(z,y) = |z - y|$ for arguments $z,y \in [0,1]$. The test error or risk is defined, as before, as 
\[
\risk(f) = \EX{\loss( f(\vx), y)},
\]
where expectation is over examples $(\vx,y)$ drawn from the unknown joint distribution from which the training set is drawn as well. 
Note that we train on the least squares loss, but we test in terms of the risk defined above. 

\subsection{Risk of early stopped neural network training}

Our main result is a bound on the test error of the two layer neural network trained for $\iter$ iterations.
The result depends on the Gram matrix $\mSigma \in \reals^{n \times n}$ determined by two kernels associated with the first and second layer of the network. The $(i,j)$-th entry of the Gram matrix as a function of the training examples is defined as
\begin{align}
\label{eq:compkernel}
\mSigma_{ij} = \nu^2
K_1(\vx_i,\vx_j)
+
\omega^2
K_2(\vx_i,\vx_j),
\end{align}
with kernels
\begin{align*}
K_1(\vx_i, \vx_j)
&=
\frac{1}{2}
\left(
1 - \frac{\cos^{-1} \left( \rho_{ij} \right)}{\pi}
\right)
\rho_{ij} \\ 
K_2(\vx_i, \vx_j)
&=
\frac{1}{2}
\left(
\frac{ \sqrt{1 - \rho_{ij}^2 } }{\pi}
+
\left(1 - \frac{\cos^{-1}(\rho_{ij})}{\pi} \right) \rho_{ij} 
\right) 
\end{align*}
where $\rho_{ij} = \innerprod{\vx_i}{\vx_j}$ (recall that we assume $\norm[2]{\vx_i} = 1$, for all $i$). 
See Figure~\ref{fig:kernels} for a plot of the kernels as a function of $\rho_{ij}$. 
Our result depends on the singular values and vectors  of this Gram matrix:
\[
\mSigma 
= 
\sum_{i=1}^n \sigma_i^2 \vu_i \transp{\vu}_i.
\]
%
%
We are now ready to state our result.

\begin{theorem}
\label{eq:informalversion}
Let $\alpha>0$ be the smallest eigenvalue of the Gram matrix $\mSigma$, suppose that the network is sufficiently wide, i.e., 
$k \geq \Omega \left( \frac{n^{10}}{\alpha^{15} \min(\nu,\omega)} \right)$, and suppose the initialization scale parameters obey $\nu \omega \leq  \alpha/ \sqrt{32\log(2n/\delta)}$ and $\nu + \omega \leq 1$ for some $\delta \in (0,1)$. 
Then, with probability at least $1-\delta$, the risk of the network trained with gradient descent for $\iter$ iterations is at most
\begin{align}
\label{eq:riskbound}
\risk(f_{\mW_\iter,\vv_\iter})
\leq
\sqrt{
\frac{1}{n}
\sum_{i=1}^n \innerprod{\vu_i}{\vy}^2 (1 - \eta \sigma_i^2)^{2\iter}
}
+
\sqrt{
\frac{1}{n}
\sum_{i=1}^n
\innerprod{\vu_i}{\vy}^2
\frac{(1 - (1 - \stepsize \sigma^2_i)^\iter)^2}{\sigma_i^2}
}
+
O(1/\sqrt{n}). 
\end{align}
\end{theorem}

Regarding the assumptions of the theorem, we remark that the exponent of $n$ and $\alpha$ in the width-condition ($k \geq \Omega \left( \frac{n^{10}}{\alpha^{15} \min(\nu,\omega)} \right)$) can be improved. Regarding the assumption that the smallest eigenvalue of the Gram matrix obeys $\alpha > 0$, Theorem 3.1 by \citet{du_GradientDescentFinds_2019} shows that if no two $\vx_i,\vx_j$ are parallel, then $\alpha>0$, for a very related Gram matrix. As argued in that work, for most real-world datasets no two inputs are parallel, therefore the assumption on the Gram matrix is rather mild.

The risk bound established by Theorem~\ref{eq:informalversion} can be interpreted as a superposition of $n$-many U-shaped bias variance tradeoffs, similar to the expression~\eqref{eq:defbarR} governing the risk of early stopped linear least squares. 
Specifically, the $i$-th ``bias'' term $\innerprod{\vu_i}{\vy}^2 (1 - \eta \sigma_i^2)^{2\iter}$ decreases in the number of gradient descent iterations $\iter$, while the $i$-th ``variance'' term $\innerprod{\vu_i}{\vy}^2
\frac{(1 - (1 - \stepsize \sigma^2_i)^\iter)^2}{\sigma_i^2}$ increases in the number of gradient descent iterations. The speed at which the two terms corresponding to the $i$-th singular value increase and decrease, respectively,  depends on the singular value $\sigma_i^2$. Those singular values, in turn, are determined by the kernels $K_1$ and $K_2$, the random initialization (in particular the scale parameters $\nu,\omega$), and the distribution of the examples. 

Whether epoch-wise double descent occurs or not depends on those singular values and therefore on the kernels, the initialization, and the distribution of the examples. 

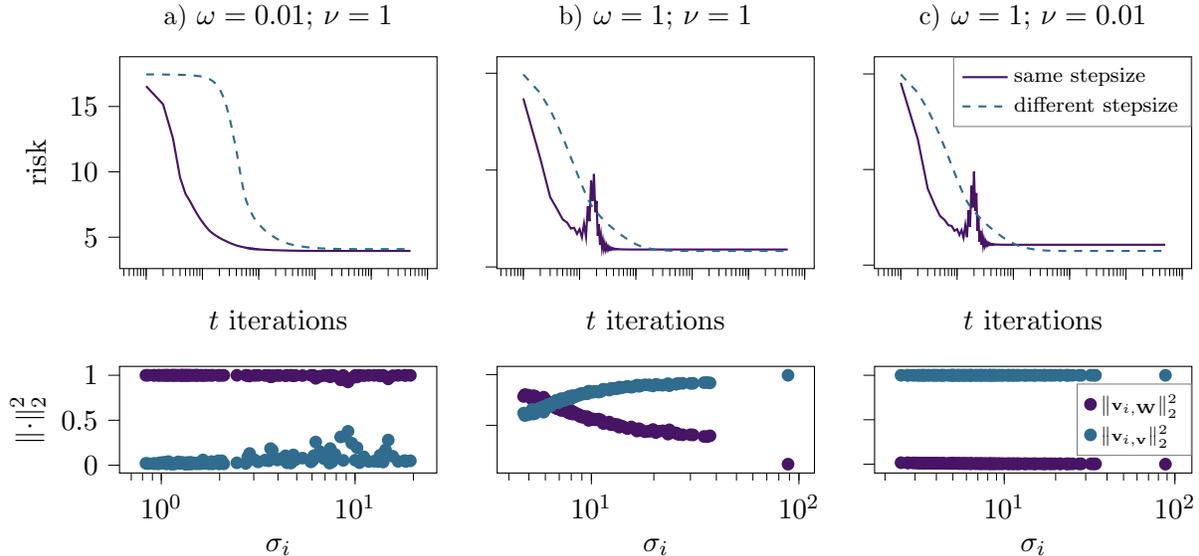
\begin{figure}[t]
\begin{center}
\begin{tikzpicture}[>=latex]

\begin{groupplot}[
width=5.8cm,
height=4.4cm,
group style={
    group name=multi_dd,
    group size= 3 by 2,
    xlabels at= all,
    ylabels at=edge left,
    yticklabels at=edge left,
    xticklabels at=edge bottom,
    horizontal sep=0.8cm, 
    vertical sep=1.3cm,
    },
legend cell align={left},
legend columns=1,
legend style={
            at={(1,1)}, 
            anchor=north east,
            draw=black!50,
            fill=none,
            font=\small
        },
colormap name=viridis,
cycle list={[colors of colormap={50,350,750}]},
tick align=outside,
tick pos=left,
x grid style={white!69.01960784313725!black},
xtick style={color=black},
y grid style={white!69.01960784313725!black},
ylabel style={at={(-0.20,0.5)}},
ytick style={color=black},
xmode=log,
title style={at={(0.5,+1.0)}},
xlabel = {$\iter$ iterations},
ylabel = {risk},
every axis plot/.append style={thick}
]

\nextgroupplot[title={{\small a)} $\omega = 0.01$; $\nu = 1$}]
      \addplot + [mark=none] table[x=t,y=dd, 
      ]{./fig/two_layer_nn_risk_s14_w001_v1.txt};
       \addplot + [mark=none,dashed] table[x=t,y=no-dd, 
       ]{./fig/two_layer_nn_risk_s14_w001_v1.txt};
       
\nextgroupplot[title={{\small b)} $\omega = 1$; $\nu = 1$}]
      \addplot + [mark=none] table[x=t,y=dd, 
      ]{./fig/two_layer_nn_risk_s14_w1_v1.txt};
       \addplot + [mark=none,dashed] table[x=t,y=no-dd, 
       ]{./fig/two_layer_nn_risk_s14_w1_v1.txt};

\nextgroupplot[title={{\small c)} $\omega = 1$; $\nu = 0.01$}]
      \addplot + [mark=none] table[x=t,y=dd, 
      ]{./fig/two_layer_nn_risk_s14_w1_v001.txt};
      \addlegendentry{\scriptsize same stepsize};
       \addplot + [mark=none,dashed] table[x=t,y=no-dd, 
       ]{./fig/two_layer_nn_risk_s14_w1_v001.txt};
       \addlegendentry{\scriptsize different stepsize};


\nextgroupplot[xlabel = {$\sigma_i$}, ylabel = {$\norm[2]{\cdot}^2$},height=3cm,]
    \addplot + [only marks] table[x=sv,y=vw]{./fig/two_layer_nn_jacobian_w001_v1.txt};
      
    \addplot + [only marks] table[x=sv,y=vv]{./fig/two_layer_nn_jacobian_w001_v1.txt};
    
    
\nextgroupplot[xlabel = {$\sigma_i$},height=3cm,]
    \addplot + [only marks] table[x=sv,y=vw]{./fig/two_layer_nn_jacobian_w1_v1.txt};
      
    \addplot + [only marks] table[x=sv,y=vv]{./fig/two_layer_nn_jacobian_w1_v1.txt};
    
    
\nextgroupplot[xlabel = {$\sigma_i$},height=3cm, 
        legend style={
            at={(1,0.5)}, 
            anchor=east,
            draw=black!50,
            fill=none,
            font=\tiny,
            cells={align=left}
        },
    ]
    \addplot + [only marks] table[x=sv,y=vw]{./fig/two_layer_nn_jacobian_w1_v001.txt};
    \addlegendentry{$\norm[2]{\vv_{i,\mW} }^2$}
      
    \addplot + [only marks] table[x=sv,y=vv]{./fig/two_layer_nn_jacobian_w1_v001.txt};
    \addlegendentry{$\norm[2]{\vv_{i,\vv}}^2$}
    

\end{groupplot}

\end{tikzpicture}
\end{center}

\vspace{-0.6cm}

\caption{
\label{fig:training_two_layer_nn}
{\bf Top row: } Risk of the two-layer neural network trained on data drawn from a linear model with diagonal covariance matrix with geometrically decaying variances. 
The weights in the first and second layer are randomly initialized with scale parameters $\omega$ and $\nu$ and the network is trained by minimizing the least-squares loss with gradient descent. 
The results show that the empirical risk computed on an independent test set has a double descent shape. Moreover, choosing a smaller stepsize for the weights in the second layer (different stepsize) eliminates the double descent and improves the risk performance as suggested by the theory.
{\bf Bottom row: } 
The norms $\norm[2]{\vv_{i,\mW} }^2$ and $\norm[2]{\vv_{i,\vv} }^2$ measure to what extend the singular values $\sigma_i$ are associated with the weights in the first ($\mW$) and second ($\vv$) layer respectively. 
Double descent occurs when singular values are mostly associated with the second layer, and therefore those weights are learned faster relative to the first layer weights.
}
\end{figure}

\paragraph{Numerical example to illustrate the theorem:}
We illustrate the above statement with a concrete numerical example. 
We draw data from the linear model specified in Section~\ref{sec:datamodel} with geometrically decaying diagonal co-variance entries $\sigma_i$ and zero additive noise, i.e., $\sigma = 0$).
We then train the network for different initialization scale parameters $\omega,\nu$ once with the same stepsize for both layers ($\eta=\text{8e-5}$), and once with a smaller stepsize for the second layer, i.e., $\eta_{\mW} = \text{8e-5}$ and $\eta_{\vv} = \text{1e-6}$. 
In the top row of Figure~\ref{fig:training_two_layer_nn}, it can be seen that the empirical risk has a double-descent behavior if both layers are initialized at the same scale (i.e., $\omega=\nu=1$) and when using the same stepsize for both layers.

To understand the relation to the theorem better, we also plot Figure~\ref{fig:training_two_layer_nn} the extent to which the singular values are associated with the parameters in the first and second layer. 
To capture this, we first comment on the relation of the parameters of the first and second layer to the singular values $\sigma_i^2$ and vectors $\vu_i^2$ of the Gram matrix. The theorem is proven by approximating the network with its linear approximation around the initialization. With this approximation, the networks predictions for the $n$ training examples are approximately given by
\begin{align}
\label{eq:jacobianmain}
\begin{bmatrix}
f_{\mW,\vv}(\vx_1) \\
\ldots \\
f_{\mW,\vv}(\vx_n) 
\end{bmatrix}
\approx
\mJ 
\begin{bmatrix}
\mathrm{Vect}(\mW)\\
\vv
\end{bmatrix}
=
\sum_{i=1}^n \sigma_i \vu_i 
( \transp{\vv} _{i,\mW} \mathrm{Vect}(\mW) 
+ 
\transp{\vv} _{i,\vv} \vv),
\end{align}
where $\mJ \in \reals^{n \times dk + k}$ is (approximately) the Jacobian of the network at initialization and $\mJ = \sum_{i=1}^n \sigma_i \vu_i \transp{\vv_i}$ is its singular value decomposition. 
Moreover, we denoted by 
$\vv _{i,\mW} \in \reals^{dk}$ and $\vv_{i,\vv} \in \reals^k$ the parts of the right-singular vectors of the Jacobian associated with the weights in the first and second layer, respectively. 
The norm of the vectors $\vv _{i,\mW}$ and $\vv _{i,\mW}$ measures to what extent the singular value $\sigma_i$ is associated with the weights in the first and second layer. 

Let us now return to the numerical example: At the bottom row of Figure~\ref{fig:training_two_layer_nn} we plot the norms of those vectors, which demonstrates that if we initialize both layers at the same scale ($\omega=\nu=1$), then most of the large singular values are associated, for the most part, with the weights in the second layer. 
This leads to double descent, that can be mitigated by choosing a smaller stepsize associated with the weights in the second layer. 

\paragraph{Improving performance by eliminating double descent:}
Similarly as for the linear least squares problem studied in the previous section, it is possible to shape the bias variance tradeoffs by adapting the stepsizes (or through initialization of the layers). However, this is not as straightforward as for the linear-least square problem with \emph{diagonal covariance matrix} studied in the previous section 
because changing the stepsize of one weight can impact all other weight updates. 

Nevertheless, we can change the stepsizes pertaining to the two different layers and thereby control whether we do or do not have double descent. 
To see why, suppose we choose the initialization equally, i.e., $\omega=1$ and $\nu=1$, but update the variables of the second layer (i.e., $\vv$) with a much larger stepsize than those of the first layer (i.e., $\mW$). Then the kernel associated with the second layer dominates and the network behaves like a random feature model~\cite{rahimi_RandomFeaturesLargeScale_2008}. 
Similarly, if we update the variables of the first layer (i.e., $\mW$) with a much larger stepsize than those of the second layer (i.e., $\mW$) then the network behaves like a network with the final layer weights $\vv$ fixed.
Thus, the stepsizes trade off the impact of the two kernels, and this tradeoff yields overlapping bias-variance tradeoffs and therefore a double descent curve.

In Figure~\ref{fig:training_two_layer_nn}, we illustrate this behavior: Double descent is eliminated by choosing a smaller stepsize for the second layer, or by choosing a smaller initialization for the first layer, 
as suggested by our theoretical results, and similar to the linear least squares setup as discussed in the previous section.
Also note that, not only does choosing a smaller stepsize for the second layer eliminate double descent, it also gives a better overall risk.

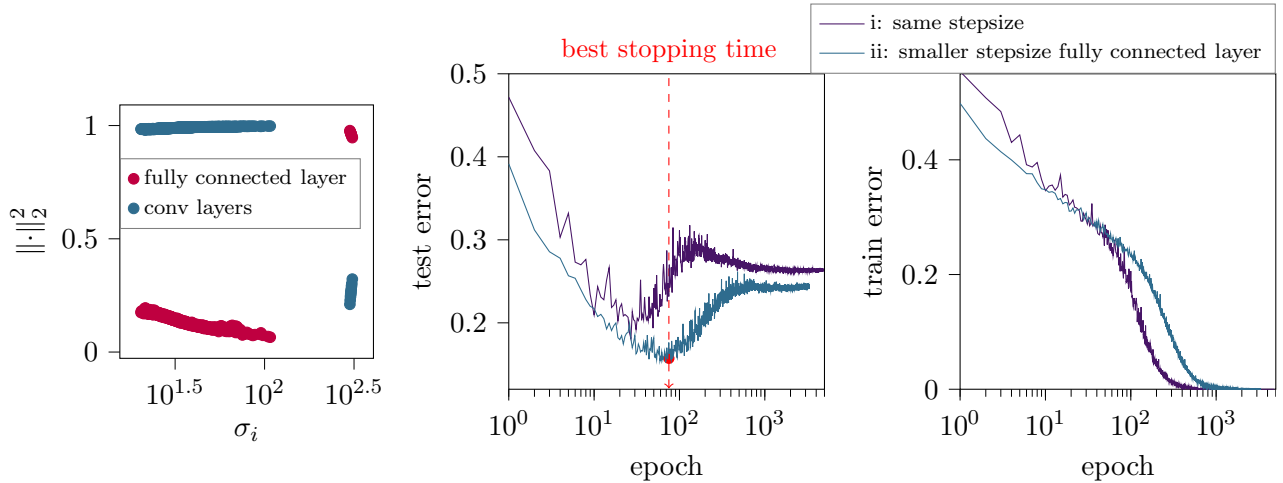
\begin{figure}[t]
\hspace{-.35cm}
\begin{tikzpicture}

\draw[red,dashed,<-] (7.3cm,-0.4cm) -- (7.3cm,3.8cm) node [above] {\small best stopping time};
\node at (7.3cm,0.0cm) [red,circle,fill,inner sep=1.5pt]{};

\begin{groupplot}[
height=0.35\textwidth,
group style={
    group name=multi_dd,
    group size= 3 by 1,
    xlabels at=edge bottom,
    ylabels at=edge left,
    xticklabels at=edge bottom,
    horizontal sep=1.8cm, 
    vertical sep=1.8cm,
    },
legend cell align={left},
legend columns=1,
legend style={
            at={(1,1)}, 
            anchor=south east,
            draw=black!50,
            fill=none,
            font=\small
        },
/tikz/every even column/.append style={column sep=0.1cm},
/tikz/every row/.append style={row sep=0.3cm},
colormap name=viridis,
cycle list={[colors of colormap={50,350,750}]},
tick align=outside,
tick pos=left,
x grid style={white!69.01960784313725!black},
xtick style={color=black},
y grid style={white!69.01960784313725!black},
ytick style={color=black},
xlabel={epoch}, 
xmode=log,
]


%
%

\nextgroupplot[xlabel = {$\sigma_i$}, 
  ylabel = {$\norm[2]{\cdot}^2$},
  width=0.3\textwidth, height = 0.3\textwidth,
        ylabel style={at={(-0.25,0.5), anchor=east}},
        legend style={
            at={(0,0.65)}, 
            anchor=west,
            draw=black!50,
            fill=none,
            font=\tiny,
            cells={align=left}
        },
    ]
    \addplot + [only marks, color=purple] table[x=sv,y=fc]{./fig/mcnn_jacobian_svd_batch.txt};
    \addlegendentry{\scriptsize fully connected layer};
      
    \addplot + [only marks] table[x=sv,y=conv]{./fig/mcnn_jacobian_svd_batch.txt};
    \addlegendentry{\scriptsize conv layers};


\nextgroupplot[ymin=0.12, ymax=0.5, ylabel={test error},width=0.35\textwidth,xmin=1,xmax=5000]

\addplot +[mark=none] table[x=epoch,y expr=1 - \thisrow{test}/100]{fig/data/double_descent_5layer.txt};

\addplot +[mark=none]
table[x=epoch,y expr=1 - \thisrow{test}/100]{fig/data/no_double_descent_5layer.txt};


\nextgroupplot[ymin=0.0, ymax=0.55, ylabel={train error},width=0.35\textwidth,xmin=1,xmax=5000]

\addplot +[mark=none] table[x=epoch,y expr=1 - \thisrow{train}/100]{fig/data/double_descent_5layer.txt};
 \addlegendentry{\scriptsize i: same stepsize};

\addplot +[mark=none]
table[x=epoch,y expr=1 - \thisrow{train}/100]{fig/data/no_double_descent_5layer.txt};
\addlegendentry{\scriptsize ii: smaller stepsize fully connected layer};


\end{groupplot}

\end{tikzpicture}

\caption{\label{fig:convnet}
Mitigating double descent for a 5-layer CNN trained on noisy CIFAR-10.
{\bf Left:} 
Norm of the parts of the right singular vectors of the Jacobian associated with the weights in the convolutional layer $\norm[2]{\vv_{i,C} }^2$ and the fully connected layer $\norm[2]{\vv_{i,F} }^2$ as a function of the singular values of the Jacobian, $\sigma_i$, showing that large singular values pertain mostly to the fully connected layer. 
If all weights have the same stepsize, this causes the weights of the fully connected layer to be learned much faster than those of the convolution layer.
{\bf Middle and Right:}
Test error of the CNN trained with the i) same stepsize for all layer, and  with ii) a smaller stepsize for the last fully connected layer.
Decreasing the learning rate of the last layer causes the last layer to be learned at a similar speed as the convolutional layers and thereby eliminates double descent and increases the best early stopped performance (i.e., the minima of ii is smaller than that of i). 
}

\end{figure}


\section{Early stopping in convolutional neural networks}
\label{sec:cnnexperiments}

In this section, we study the training of a standard  5-layer convolutional neural network (CNN) and a standard ResNet-18 model on the (10 class classification) CIFAR-10 dataset. Both networks were studied in~\cite{nakkiran_DeepDoubleDescent_2020}, and we note that, as shown in that paper, the risk (or test error) only has a double descent behavior if the network is trained on a dataset with label noise. 
In this section we consider the same setup and train on the CIFAR-10 training set with 20\% random label noise. 
While we have no theoretical results for those two complicated neural network models, we demonstrate---inspired by our theory---that epoch-wise double descent can be eliminated and the early stopping performance can be improved by adjusting the stepsizes/learning rates.

\paragraph{5-layer CNN:} 
The 5-layer CNN consists for 4 convolutional layers followed by one fully connected layer. We demonstrate that, just like for the two-layer network from the previous section, double descent can be eliminated by changing the stepsize of the final layer, this time by decreasing the stepsize of the final fully connected layer. 

Figure~\ref{fig:convnet} depicts the test error of the network trained on noisy CIFAR-10 with i) the same stepsize for all layers, and with ii) a smaller stepsize for the fully connected layer, along with the corresponding training error curves.
The figure shows that in the standard setting, i) double descent occurs, whereas ii) choosing a smaller stepsize for the fully connected layer eliminates double descent.

To see the intuition behind this approach, let us consider the Jacobian of the
5-layer CNN similarly as we considered the Jacobian of the two layer network given by~\eqref{eq:jacobianmain}
in the previous section.
We again compute the Jacobian of the networks' predictions at initialization and plot the norm of the part of the right singular vectors $\vv_{i,C}$ and $\vv_{i,F}$ associated with the weights in the convolutional and the fully connected layers, respectively. 
The plot shows that large singular values are associated mostly with the weights of the fully connected layers. This causes the convolutional layers to be learned slower than the fully connected layer which results in double descent. 
Analogously as in the previous section, decreasing the stepsize pertaining to the fully connected layer eliminates double descent. 

\begin{figure}
  \begin{center}
  \begin{tikzpicture}[>=latex]
  
  
  
  
  \begin{groupplot}[
  width=8cm,
  height=5cm,
  group style={
      group name=resnet_dd,
      group size= 2 by 1,
      xlabels at=edge bottom,
      ylabels at=edge left,
      xticklabels at=edge bottom,
      horizontal sep=1.8cm, 
      vertical sep=1.8cm,
      },
  legend cell align={left},
  legend columns=1,
  legend style={
              at={(1,1)}, 
              anchor=north east,
              draw=black!50,
              fill=none,
              font=\small,
          },
  colormap name=viridis,
  cycle list={[colors of colormap={50,350,750}]},
  tick align=outside,
  tick pos=left,
  x grid style={white!69.01960784313725!black},
  xtick style={color=black},
  y grid style={white!69.01960784313725!black},
  ytick style={color=black},
  xlabel={epoch}, 
  xmode=log,
  ]
  
  
  \nextgroupplot[ymin=0.12, ymax=0.55, ylabel={test error}]
  
  \addplot +[mark=none] table[x=epoch,y expr=1 - \thisrow{test}/100, select coords between index={0}{1999}]{fig/data/resnet_double_descent.txt};
  \addlegendentry{\scriptsize i: same stepsize};
  \addplot +[mark=none] table[x=epoch,y expr=1 - \thisrow{test}/100]{fig/resnet_scaled_lr_no_dd.txt};
  \addlegendentry{\scriptsize ii: smaller stepsize last half layers};
  
  \nextgroupplot[ymin=0, ymax=0.42, ylabel={train error}]
  
  \addplot +[mark=none] table[x=epoch,y expr=1 - \thisrow{train}/100, select coords between index={0}{1999}]{fig/data/resnet_double_descent.txt};
  \addplot +[mark=none] table[x=epoch,y expr=1 - \thisrow{train}/100]{fig/resnet_scaled_lr_no_dd.txt};
  
  \end{groupplot}
  \end{tikzpicture}  
  \end{center}
  
  \vspace{-0.6cm}
  
  \caption{
  \label{fig:doubledescentresnet}
  {\bf Left:} Test error of the ResNet-18 trained with the i) same stepsize for all
  layer, and  with ii) a smaller stepsize for the latter half of the layers.
  Decreasing the learning rate of the last layers causes the last layers to be
  learned at a similar speed as the first and thereby eliminates
  double descent. 
  {\bf Right:} The training error curves for i) and ii).
  }
\end{figure}
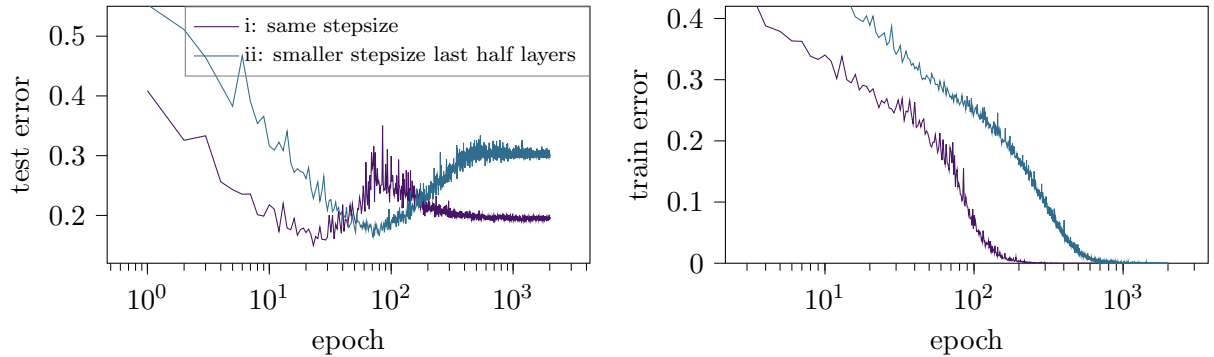

\paragraph{ResNet-18:} 
We next consider a more complex and deeper network, namely the popular ResNet-18 model. 
ResNet-18 has a double descent behavior when trained on the previously considered CIFAR-10
problem with 20\% random label noise~\cite{nakkiran_DeepDoubleDescent_2020}. 
Inspired by our theory, we again hypothesize that the double descent behavior occurs because some layer(s) of the ResNet-18 model are fitted at a faster rate than others. If that hypothesis is true, then scaling the learning rates of some layers should eliminate double descent. Indeed, Figure~\ref{fig:doubledescentresnet} shows that when scaling the stepsizes of the later half of the layers of the network mitigates double descent. 


\section*{Code}
Code to reproduce the experiments is available at 
\url{https://github.com/MLI-lab/early_stopping_double_descent}.

\section*{Acknowledgements}
R. Heckel and F. F. Yilmaz are (partially) supported by NSF award IIS-1816986. 
R. Heckel acknowledges support of the NVIDIA Corporation in form of a GPU, and would like to thank Fanny Yang and Alexandru Tifrea for discussions and helpful comments on this manuscript.  

\printbibliography

\appendix

\section{Numerical results for linear least squares}

In this section we provide further numerical results for linear least squares.
We consider a linear model with $d=700$ features, and with $n=6d$ examples. 
We let a fraction $6/7$ of the features have singular value $\sigma_i = 1$ and associated model coefficient $\theta_i = 1$, and the rest, $1/7$ of the features, have singular value $\sigma_i=0.1$ and $\theta_i =10$. 
In Figure~\ref{fig:doubledls}(a) we show the risk obtained by simulating the risk empirically along with the risk expression $\bar R(\tilde \vtheta^\iter)$ given by equation~\eqref{eq:defbarR}. It can be seen that the risk expression $\bar R(\tilde \vtheta^\iter)$ slightly under-estimates the true risk. 
The quality of the estimate becomes better as we increase $n$; in Figure~\ref{fig:doubledls}(b) we show simulations for the same configuration but with $n=10d$. 


\newcommand{\errorband}[6]{ \pgfplotstableread{#1}\datatable \addplot
  [name path=pluserror,draw=none,no markers,forget plot] table
  [x={#2},y expr=\thisrow{#3}+\thisrow{#4}] {\datatable};

  \addplot [name path=minuserror,draw=none,no markers,forget plot]
    table [x={#2},y expr=\thisrow{#3}-\thisrow{#4}] {\datatable};

  \addplot [forget plot,fill=#5,opacity=#6]
    fill between[on layer={},of=pluserror and minuserror];

  \addplot [#5,thick,no markers]
    table [x={#2},y={#3}] {\datatable};
}

\begin{figure}[t]
\begin{center}
\begin{tikzpicture}
\begin{groupplot}[
         title style={at={(0.5,-0.35)},anchor=north}, group
         style={group size=2 by 1, horizontal sep=2cm },
         width=0.4\textwidth, height=0.34\textwidth, xticklabel style={/pgf/number
           format/fixed, /pgf/number format/precision=3}, yticklabel
         style={/pgf/number format/fixed, /pgf/number
           format/precision=5}, xmode=log] 

\nextgroupplot[xlabel = {$\iter$ iterations},ylabel = {risk},title={(a) $n=5d$}]
      
      \errorband{./fig/risk_n=5d.dat}{0}{2}{3}{blue}{0.2}
      \addlegendentry{$\risk(\vtheta^\iter)$};
      \addplot + [mark=none] table[x index=0,y index=1]{./fig/risk_n=5d.dat};
      \addlegendentry{$\bar R(\tilde \vtheta^\iter)$};

\nextgroupplot[xlabel = {$\iter$ iterations},title={(b) $n=10d$} ]
      
      \errorband{./fig/risk_n=10d.dat}{0}{2}{3}{blue}{0.2}
      \addlegendentry{$\risk(\vtheta^\iter)$};

      \addplot + [mark=none] table[x index=0,y index=1]{./fig/risk_n=10d.dat};
      \addlegendentry{$\bar R(\tilde \vtheta^\iter)$};
      
\end{groupplot}          
\end{tikzpicture}
\end{center}

\vspace{-0.6cm}

\caption{
The risk of early-stopped gradient least-squares $\risk (\tilde \vtheta^\iter)$ based on numerical simulation of the Gaussian model along with the risk expression $\bar R(\tilde \vtheta^\iter)$ given in~\eqref{eq:defbarR}. 
We averaged over 100 runs of gradient descent, and the shaded region corresponds to one standard deviation over the runs. 
It can be seen that the risk expression slightly underestimates the true risk, but other than that describes the behavior of the risk well.
\label{fig:doubledls}
}
\end{figure}
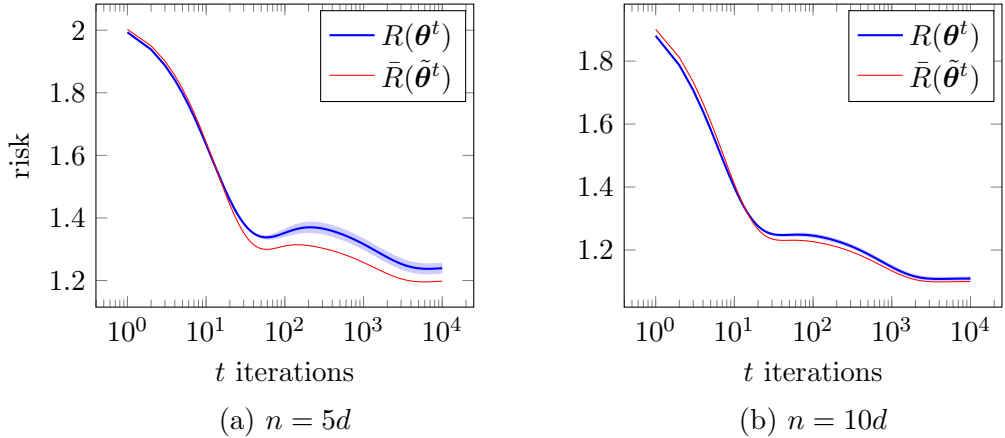


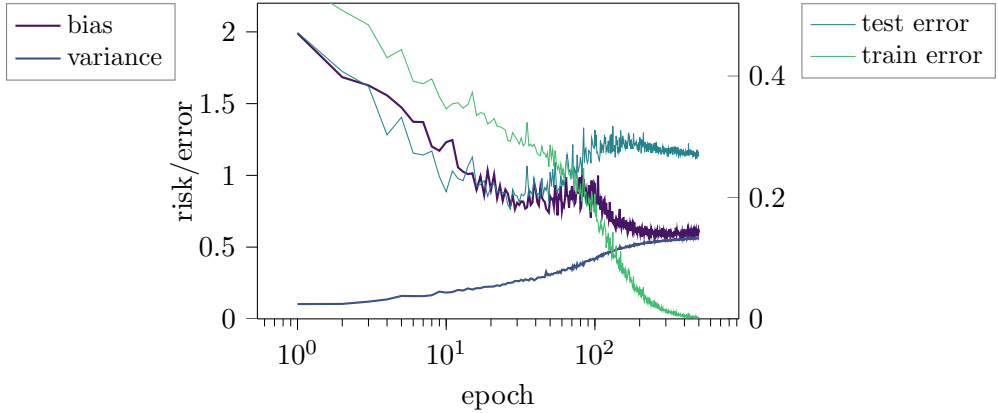
\begin{figure}[t]
\begin{center}
\pgfplotsset{set layers}%
\begin{tikzpicture}[>=latex]
\begin{axis}[
height=4.2cm,
width=6.4cm,
ymin=0, ymax=2.2,
scale only axis,
axis y line*=left,
legend cell align={left},
legend columns=1,
legend style={
            at={(-0.17,1)}, 
            anchor=north east,
            draw=black!50,
            fill=none,
            font=\small
        },
/tikz/every even column/.append style={column sep=0.1cm},
/tikz/every row/.append style={row sep=0.3cm},
colormap name=viridis,
cycle list={[colors of colormap={50,250,450,700}]},
tick align=outside,
tick pos=left,
x grid style={white!69.01960784313725!black},
xtick style={color=black},
y grid style={white!69.01960784313725!black},
ytick style={color=black},
xlabel={epoch}, 
ylabel={risk/error},
ylabel style = {at={(-0.1,0.5)}, align=center},
xmode=log,
every axis plot/.append style={thick}
]

\addplot +[mark=none] table[x=epoch,y=bias]{fig/mcnn_bias_variance_smooth.txt};
 \addlegendentry{bias};

\addplot +[mark=none]
table[x=epoch,y=variance]{fig/mcnn_bias_variance_smooth.txt};
\addlegendentry{variance};
\end{axis}
\begin{axis}[
height=4.2cm,
width=6.4cm,
scale only axis,
ymin=0, ymax=0.52,
axis y line*=right,
axis x line=none,
legend cell align={left},
legend columns=1,
legend style={
            at={(1.13,1)}, 
            anchor=north west,
            draw=black!50,
            fill=none,
            font=\small
        },
xmode=log,
colormap name=viridis,
cycle list={[colors of colormap={450,700}]},
]
\addplot +[mark=none] table[x=epoch,y expr=1 - \thisrow{test}/100, select coords between index={0}{499}]{fig/data/double_descent_5layer.txt};
\addlegendentry{test error};

\addplot +[mark=none] table[x=epoch,y expr=1 - \thisrow{train}/100, select coords between index={0}{499}]{fig/data/double_descent_5layer.txt};
\addlegendentry{train error};
\end{axis}
\end{tikzpicture}
\end{center}
\vspace{-0.6cm}
\caption{
\label{fig:bias_variance}
Bias and variance as a function on training epochs for training a 5-layer CNN until convergence.
The overall variance is increasing, and the overall bias has a double-descent like shape.
The training and test error curves show the interpolation of the training set and double descent behavior of the error in this interval.
}
\end{figure}

\section{Numerical bias-variance decomposition for the 5-layer CNN}
\label{sec:bias-variancedecomposition}

As discussed before, classical machine learning theory for the underparameterized regime establishes the bias-variance tradeoff as a result of the bias decreasing and the variance increasing as a function of the model size (complexity). In the over-parameterized regime, the bias often continues to decrease, while the variance also decreases. This has been established in a number of recent works~\cite{jacot_ImplicitRegularizationRandom_2020,yang_RethinkingBiasVarianceTradeoff_2020,dascoli_DoubleTroubleDouble_2020}, and provides a bias-variance decomposition of the model-wise double-descent shaped risk curve.

In this paper, we showed that the epoch-wise double descent occurs for a different reason than the model-wise double descent. Namely, epoch-wise double descent can be explained as a temporal superposition of multiple bias-variance tradeoff curves rather than with a unimodal variance curve.

That also means that the overall bias (i.e., the sum of the individual bias terms) might not be decreasing and the overall variance might not be uni-modal like in the model-wise case. To demonstrate that it is in fact not, in Figure~\ref{fig:bias_variance}, we plot the numerically computed bias and variance terms (computed as proposed
in~\cite{yang_RethinkingBiasVarianceTradeoff_2020}) for the CNN experiment from Section~\ref{sec:cnnexperiments} along with the risk, which shows that in fact the overall bias is increasing, while the variance has a double-descent like shape.


\section{Proof of Theorem~\ref{thm:maindifference}}

The difference of the risk and risk expression can be bounded by
\begin{align}
\label{eq:stdbound}
\left| \risk(\vtheta^\iter) - \bar R(\tilde \vtheta^\iter) \right|
&\leq
\left| \risk(\vtheta^\iter) - \risk(\tilde \vtheta^\iter) \right|
+
\left| 
\risk(\tilde \vtheta^\iter) - \bar R(\tilde \vtheta^\iter)
\right|.
\end{align}
We bound the two terms on the righ-hand-side separately. We start with bounding the first term by applying the lemma below.

\begin{lemma}
\label{lem:bounditerates}
Define $\tilde \mX$ so that $\mX = \tilde \mX \mSigma$.  
Suppose that $\norm{ \mI - \transp{\tilde \mX} \tilde \mX} \leq \epsilon$, with $\epsilon \leq \frac{\min_i \eta_i \sigma_{i}^2}{2 \max_i \eta_i \sigma_{i}^2}$. Then
\begin{align}
\label{eq:bounditerates}
\left| \risk(\vtheta^\iter) - \risk(\tilde \vtheta^\iter) \right|
\leq
(1 - (1 - \min_i \eta_i \sigma_{i}^2 / 2)^\iter)^2
8 \frac{\max_i \eta_i^2 \sigma_{i}^4}{\min_i \eta_i^2 \sigma_{i}^4} \epsilon^2 \max_{\ell \in \{1,\ldots, k\}} \norm[2]{\mSigma \tilde \vtheta^{\ell} - \mSigma  \vtheta^\ast}^2.
\end{align}
\end{lemma}

In order to apply the lemma, we start by verifying its condition. Towards this goal, consider the matrix $\mX = \tilde \mX \mSigma$ and note that the entries of $\tilde \mX$ are iid $\mc N(0,1/n)$.
A standard concentration inequality from the compressive sensing literature (specifically~\cite[Chapter 9]{foucart_MathematicalIntroductionCompressive_2013}) states that, for any $\beta \in (0,1)$,
\begin{align*}
\PR{ \norm{ \mI - \transp{\tilde \mX} \tilde \mX } \geq \beta }
\leq e^{- \frac{n\beta^2}{15} + 4d}.
\end{align*}
With $\beta  = \sqrt{\frac{75d}{n}}$ we obtain that, with probability at least $1-e^{-d}$, 
\begin{align*}
 \norm{ \mI - \transp{\tilde \mX} \tilde \mX } \leq \sqrt{75\frac{d}{n}}.
\end{align*}
Next, we bound the term on the RHS of in~\eqref{eq:bounditerates}, with the following lemma.

\begin{lemma}
\label{lem:boundmax}
Provided that $\eta_i \sigma_i^2 \leq 1$ for all $i$, with probability at least $1 - 2d(e^{-\beta^2/2} + e^{-n/8})$, 
\begin{align*}
\max_\ell
\norm[2]{\mSigma \tilde \vtheta^{\ell} - \mSigma \vtheta^{\ast} }^2
\leq
2 \norm[2]{\mSigma \vtheta^{\ast} }^2 
+
4 \frac{d}{n} \sigma^2 \beta^2.
\end{align*}
\end{lemma}

Applying the lemma with $\beta^2 = 10\log(d)$, we obtain that with probability at least
$1 - 2d^{-5} + 2de^{-n/8} - e^{-d}$ we have
\begin{align}
\label{eq:bounditerates2}
\left| \risk(\vtheta^\iter) - \risk(\tilde \vtheta^\iter) \right|
\leq
8 \frac{\max_i \eta_i^2\sigma_{i}^4}{\min_i \eta_i^2 \sigma_{i}^4} 
\frac{75d}{n}
\left(
2 \norm[2]{\mSigma \vtheta^{\ast} }^2 
+
4 \frac{d}{n} \sigma^2 10 \log(2d)
\right).
\end{align}
We are now ready to bound the second term in~\eqref{eq:stdbound}:
\begin{lemma}
\label{lem:riskvsexp}
With probability at least $1 - 4e^{- \frac{\beta^2}{8}}$, we have that
\begin{align}
\label{eq:riskvsexp}
\left| 
\risk(\tilde \vtheta^\iter) - \bar R(\tilde \vtheta^\iter)
\right|
\leq
\frac{\sigma^2}{n} \beta
3 \sqrt{d},
\end{align}
with $\bar R(\tilde \vtheta^\iter)$ as defined in~\eqref{eq:defbarR}. 
\end{lemma}

Applying the two bounds~\eqref{eq:bounditerates2} and~\eqref{eq:riskvsexp} to the RHS of the bound~\eqref{eq:stdbound} concludes the proof. 
The remainder of the proof is devoted to proving the three lemmas above.


\subsection{Proof of Lemma~\ref{lem:bounditerates}}

Recall that the iterates of the original and closely related problem are given by
\begin{align*}
\vtheta^{\iter+1} - \vtheta^\ast 
&=
(\mI - \mEta \transp{\mX} \mX) (\vtheta^{\iter} - \vtheta^\ast )
+
\mEta \transp{\mX}\vz,
 \\
\tilde \vtheta^{\iter+1} - \vtheta^\ast 
&=
\left( \mI - \mEta \transp{\mSigma} \mSigma \right) (\tilde \vtheta^{\iter} - \vtheta^\ast)
+
\mEta \transp{\mX}\vz.
\end{align*}
Note that $\mX = \tilde \mX \mSigma$, where we defined $\tilde \mX$ which has iid Gaussian entries $\mc N(0,1/n)$.
With this notation, and using that $\mSigma$ is diagonal and therefore commutes with diagonal matrices, we obtain the following expressions for the residuals of the two iterates:
\begin{align*}
\mSigma \vtheta^{\iter+1} - \mSigma \vtheta^\ast 
&=
(\mI - \mEta \mSigma^2 \transp{\tilde \mX} \tilde \mX ) (\mSigma\vtheta^{\iter} - \mSigma \vtheta^\ast )
+
\mEta \mSigma^2 \transp{\tilde \mX}\vz
 \\
\mSigma \tilde \vtheta^{\iter+1} - \mSigma \vtheta^\ast 
&=
\left( \mI - \mEta \mSigma^2 \right) (\mSigma \tilde \vtheta^{\iter} - \mSigma  \vtheta^\ast)
+
\mEta \mSigma^2 \transp{\tilde \mX}\vz.
\end{align*}
The difference between the residuals is
\begin{align*}
\mSigma \vtheta^{\iter+1} 
-
\mSigma \tilde \vtheta^{\iter+1}
&= 
(\mI - \mEta \mSigma^2 \transp{\tilde \mX} \tilde \mX ) (\mSigma\vtheta^{\iter} - \mSigma \vtheta^\ast )
-
\left( \mI - \mEta \mSigma^2 \right) (\mSigma \tilde \vtheta^{\iter} - \mSigma  \vtheta^\ast) \\
&= 
\mSigma\vtheta^{\iter} - \mSigma \tilde \vtheta^{\iter}
-
\mEta \mSigma^2 \transp{\tilde \mX} \tilde \mX  (\mSigma\vtheta^{\iter} - \mSigma \vtheta^\ast )
+ 
\mEta \mSigma^2 
(\mSigma \tilde \vtheta^{\iter} - \mSigma  \vtheta^\ast) \\
&= 
(\mI - \mEta \mSigma^2 \transp{\tilde \mX} \tilde \mX)
(\mSigma\vtheta^{\iter} - \mSigma \tilde \vtheta^{\iter})
+
\mEta \mSigma^2 
(\mI - \transp{\tilde \mX} \tilde \mX)
(\mSigma \tilde \vtheta^{\iter} - \mSigma  \vtheta^\ast),
\end{align*}
where the last equality follows by adding and subtracting $\mEta \mSigma^2 \transp{\tilde \mX} \tilde \mX(\mSigma \tilde \vtheta^{\iter} - \mSigma  \vtheta^\ast)$ and rearranging the terms.
It follows that
\begin{align}
\label{eq:boundforit}
\norm[2]{\mSigma \vtheta^{\iter+1} 
-
\mSigma \tilde \vtheta^{\iter+1}}
&\leq
(1 - \min_i \eta_i \sigma_{i}^2/2)
\norm[2]{\mSigma\vtheta^{\iter} - \mSigma \tilde \vtheta^{\iter}}
+
\max_i \eta_i \sigma_{i}^2
\epsilon
\max_\ell \norm[2]{\mSigma \tilde \vtheta^{\ell} - \mSigma  \vtheta^\ast}.
\end{align}
Here, we used the bound 
\begin{align*}
\norm{\mI - \mEta \mSigma^2 \transp{\tilde \mX} \tilde \mX}
&\leq
\norm{\mI - \mEta \mSigma^2}
+
\norm{\mEta \mSigma^2 (\mI - \transp{\tilde \mX} \tilde \mX)} \\
&\leq
(1 -  \min_i \eta_i \sigma_i^2)
+
\max_i \eta_i \sigma_{i}^2 \epsilon \\
&\leq
(1 - \min_i \eta_i \sigma_{i}^2/2).
\end{align*}
Here, we used that $\eta_i \sigma_i^2 \leq 1$, by assumption, and the last inequality follows by the assumption $\epsilon \leq \frac{\min_i \eta_i \sigma_{i}^2}{2\max \eta_i \sigma_{i}^2}$.
Iterating the bound~\eqref{eq:boundforit} yields 
\begin{align*}
\norm[2]{\mSigma \vtheta^{\iter} - \mSigma \tilde \vtheta^{\iter}}
\leq
\frac{1 - (1- \min_i \eta_i \sigma_{i}^2/2)^\iter}{\min_i \eta_i \sigma_{i}^2/2}
\max_i \eta_i \sigma^2_{i} \epsilon \max_\ell \norm[2]{\mSigma \tilde \vtheta^{\ell} - \mSigma  \vtheta^\ast},
\end{align*}
which concludes the proof.

\subsection{Proof of Lemma~\ref{lem:boundmax}}
Recall that 
\[
\sigma_i(\tilde \theta^\iter_i - \theta^\ast_i)
= 
\sigma_i(1 - \eta_i \sigma_i^2)^\iter \theta^\ast_i + \transp{\tilde \vx}_i \vz (1 - (1 - \eta_i \sigma_i^2)^\iter).
\]
With $\eta_i \sigma_i^2 \leq 1$, by assumption, it follows that
\begin{align}
\label{eq:sigmaidifftheta}
\sigma_i^2(\tilde \theta^\iter_i - \theta^\ast_i)^2
&\leq
2\sigma_i^2(\theta^\ast_i)^2
+
2(\transp{\tilde \vx}_i \vz)^2.
\end{align}
Conditioned on $\vz$, the random variable $\transp{\tilde \vx}_i \vz$ is zero-mean Gaussian with variance $\norm[2]{\vz}/n$. Thus,
$\PR{ |\transp{\tilde \vx}_i \vz|^2 \geq \frac{\norm[2]{\vz}^2}{n}\beta^2} \leq 2 e^{-\beta^2/2}$. 
Moreover, as used previously in~\eqref{eq:event2}, with probability at least $1-2e^{-n/8}$, $\norm[2]{\vz}^2 \leq 2 \sigma^2$. 
Combining the two with the union bound, we obtain
\[
\PR{ |\transp{\tilde \vx}_i \vz|^2 \geq \frac{2\sigma^2}{n}\beta^2} \leq 2 e^{-\beta^2/2} + 2 e^{-n/8}.
\]
Using this bound in inequality~\eqref{eq:sigmaidifftheta}, we have that, with probability at least $1-2( e^{-\beta^2/2} + e^{-n/8})$ that
\begin{align*}
\sigma_i^2(\tilde \theta^\iter_i - \theta^\ast_i)^2
&\leq
2\sigma_i^2(\theta^\ast_i)^2
+
4\frac{1}{n} \sigma^2 \beta^2.
\end{align*}
By the union bound over all $i$ we therefore get that
\begin{align*}
\max_\iter
\norm[2]{\mSigma \tilde \vtheta^{\iter} - \mSigma \vtheta^{\ast} }^2
\leq
2 \norm[2]{\mSigma \vtheta^{\ast} }^2 
+
4 \frac{d}{n} \sigma^2 \beta^2,
\end{align*}
with probability at least $1 - 2d(e^{-\beta^2/2} + e^{-n/8})$.

\subsection{Proof of Lemma~\ref{lem:riskvsexp}}

We have
\begin{align*}
\risk(\tilde \vtheta^\iter)
&=
\sigma^2 +
\sum_{i=1}^d \sigma_i^2 
\left(
(1 - \eta_i \sigma_i^2)^\iter \theta^\ast_i + \sigma_i \transp{\tilde \vx}_i\vz \frac{1 - (1-\eta_i \sigma_i^2)^\iter}{\sigma_i^2}
\right)^2\\
&=
\sigma^2 +
\sum_{i=1}^d  
\underbrace{\left(
\sigma_i(1 - \eta_i \sigma_i^2)^\iter \theta^\ast_i +  \transp{\tilde \vx}_i\vz (1 - (1-\eta_i \sigma_i^2)^\iter
\right)^2}_{Z_i}.
%
\end{align*}
The random variable $Z_i$, conditioned on $\vz$, is a squared Gaussian with variance 
upper bounded by $\frac{\norm[2]{\vz}}{\sqrt{n}}$ and has expectation 
\[
\EX{Z_i}
=
\sigma_i^2
(1 - \eta_i \sigma_i^2)^{2\iter} (\theta^\ast_i)^2 + 
\frac{ \norm[2]{\vz}^2 }{n} (1 - (1-\eta_i \sigma_i^2)^{\iter})^2.
\]
By a standard concentration inequality of sub-exponential random variables (see e.g.~\cite[Chapter 2, Equation~2.21]{wainwright_HighDimensionalStatistics_2019}), 
we get, for $\beta \in (0,\sqrt{d})$ and conditioned on $\vz$, that the event
\begin{align}
\label{eq:event1}
\mc E_1
=
\left\{
\left| \sum_{i=1}^d (Z_i - \EX{Z_i}) \right|
\leq
\frac{\norm[2]{\vz}^2}{n} \sqrt{d} \beta
\right\}
\end{align}
occurs with probability at least $1-2e^{ -\frac{\beta^2}{8} }$.
With the same standard concentration inequality for sub-exponential random variables, we have that the event
\begin{align}
\label{eq:event2}
\mc E_2 = 
\left\{
\left| \norm[2]{\vz}^2 - \sigma^2 \right| \leq \frac{\sigma^2 \beta}{\sqrt{n}}
\right\}
\end{align}
also occurs with probability at least $1-2e^{ -\frac{\beta^2}{8} }$. 
By the union bound, both events hold simultaneously with probability at least $1-4e^{ -\frac{\beta^2}{8} }$. 
On both events, we have that
\begin{align*}
\left| 
\risk(\tilde \vtheta^\iter) - \bar R(\tilde \vtheta^\iter)
\right|
&=
\left|
\sum_{i=1}^d (Z_i - \EX{Z_i})
+
\frac{1}{n} \left( \norm[2]{\vz}^2 - \sigma^2 \right)(1 - (1-\eta \sigma_i^2)^\iter )^2
\right| \\
&\leq
\left|
\sum_{i=1}^d (Z_i - \EX{Z_i})
\right|
+
d\left| \norm[2]{\vz}^2 - \sigma^2 \right| 
\\
&\leq
\frac{\norm[2]{\vz}^2}{n} \sqrt{d} \beta 
+
\frac{d}{n} \frac{1}{\sqrt{n}} \sigma^2 \beta 
\\
&\leq
\frac{2\sigma^2}{n} \sqrt{d} \beta 
+
\frac{d}{n} \frac{1}{\sqrt{n}} \sigma^2 \beta 
\\
&\leq
\frac{\sigma^2}{n} \beta
3 \sqrt{d}.
\end{align*}
concluding the proof of our lemma.


\section{Proof of Proposition~\ref{prop:mincond}}

By equation~\eqref{eq:defbarR}, the risk expression is a sum of U-shaped curves: $R(\tilde \vtheta^\iter) =  \sigma^2 + \sum_{i=1}^d U_i(\iter)$.
We start by considering one such U-shaped curve, and find its minimum as a function of the number of iterations, $\iter$. Towards this end, we set the derivative of one such U-shaped curve, given by
\begin{align*}
\frac{\partial }{\partial k} U_{i}(\iter)
&=
\sigma_i^2
(\theta^\ast_i)^2
2\log(1 - \eta_i \sigma_i^2)
(1 - \eta_i \sigma_i^2)^{2\iter}  
+ 
\frac{\sigma^2}{n} 
2((1-\eta_i \sigma_i^2)^\iter  -  1)
\log(1 - \eta_i \sigma_i^2) (1 - \eta_i \sigma_i^2)^\iter \\
&=
2\log(1 - \eta_i \sigma_i^2)
(1 - \eta \sigma_i^2)^{\iter}  
\left(
(1 - \eta_i \sigma_i^2)^{\iter}
(\sigma_i^2
(\theta^\ast_i)^2 + \frac{\sigma^2}{n} )
-
\frac{\sigma^2}{n}
\right)
\end{align*}
to zero, which gives that the minimum occurs when 
\begin{align}
\label{eq:stepsize}
\eta_i = \frac{1}{\sigma_i^2} \left( 1 - \left( \frac{\sigma^2 /n }{ \sigma_i^2 (\theta_i^\ast)^2 + \sigma^2 /n }  \right)^{1/\iter} \right).
\end{align}
For the iteration $\iter$ which satisfies this equation, we get
\[
\min_\iter U_{i}(\iter) = \frac{ \sigma^2 /n \sigma_i^2 (\theta_i^\ast)^2 }{\sigma^2 /n + \sigma_i^2 (\theta_i^\ast)^2},
\]
thus this minimum is independent of the iteration $\iter$ and independent of the stepsize, provided their relation is as described in~\eqref{eq:stepsize} above. 



\section{Proof and statements for neural networks}

In this section, we prove the following result, which is a slightly more formal version of our main result for neural networks, Theorem~\ref{eq:informalversion}.

\begin{theorem}
\label{thm:mainthm}
Draw a dataset $\setD = \{(\vx_1,y_1),\ldots, (\vx_n,y_n)\}$ consisting of $n$ examples i.i.d. from a distribution with $\norm[2]{\vx_i} = 1$ and $|y_i| \leq 1$. 
Let $\mSigma \in \reals^{n\times n}$ be the corresponding Gram matrix defined in~\eqref{eq:compkernel}, and suppose its smallest singular value obeys $\alpha > 0$. 

Pick an error parameter $\xi \in (0,1)$ and a failure probability $\delta \in (0,1)$, and consider the two-layer neural network 
$f_{\mW,\vv}(\vx) = \frac{1}{\sqrt{k}} \relu(\transp{\vx} \mW) \vv$, with parameters $\mW^{d\times k}, \vv \in \reals^k$ initialized according to~\eqref{eq:initialization} with initialization scale parameters $\nu,\omega$ obeying
$\nu \omega \leq  \xi / \sqrt{32\log(2n/\delta)}$ and $\nu + \omega \leq 1$. 
Suppose that the network is sufficiently overparameterized, i.e., 
\begin{align}
\label{eq:condkmainthm}
k \geq \Omega \left( 
\frac{n^{10} }{\alpha^{11} \min(\nu,\omega) \xi^4}
\right).
\end{align}
Then, the risk of the network trained with gradient descent with constant stepsize $\eta$ for $\iter$ iterations obeys, with probability at least $1-\delta$,
\begin{align}
\label{eq:finalriskbound}
\risk(f_{\mW_\iter, \vv_\iter})
\leq
\sqrt{
\frac{1}{n}
\sum_{i=1}^n \innerprod{\vu_i}{\vy}^2 (1 - \eta \sigma_i^2)^{2\iter}
}
+ 
\sqrt{
\frac{1}{n}
\sum_{i=1}^n
\innerprod{\vu_i}{\vy}^2
\frac{1 - (1 - \stepsize \sigma^2_i)^{2\iter}}{\sigma_i^2}
}
+
\frac{1}{\sqrt{n}} 
+
O(\xi/\alpha).
\end{align}

\end{theorem}

Theorem~\ref{eq:informalversion} directly follows by choosing the error parameter as $\xi = O(\alpha)$.

\subsection{Proof of Theorem~\ref{thm:mainthm}}

In this section, we provide a proof of Theorem~\ref{eq:informalversion}. 
Our proof relies on the observation that highly overparameterized neural networks behave as associated linear models, as established in a large number of prior works~\citep{arora_FineGrainedAnalysisOptimization_2019,du_GradientDescentProvably_2018b,
oymak_ModerateOverparameterizationGlobal_2020,
oymak_GeneralizationGuaranteesNeural_2019,heckel_DenoisingRegularizationExploiting_2020}. 

The proof consists of two parts. 
First, we control the empirical risk as a function of the number of gradient descent steps, $\iter$.
Second, we control the generalization error, i.e., the gap between the population risk and the empirical risk by bounding the Rademacher complexity of the function class consisting of two-layer networks trained with $\iter$ iterations of gradient descent.
Recall that our result depends on the singular values and vectors of the gram matrix of kernels associated with the two-layer network. The Gram matrix is given as the expectation of the outer product of the Jacobian of the network at initialization:
\[
\mSigma 
=
\EX{ \Jac(\mW_0,\vv_0) \transp{\Jac}(\mW_0,\vv_0) }
= 
\sum_{i=1}^n \sigma_i^2 \vu_i \transp{\vu}_i.
\]
Here, expectation is with respect to the random initialization $\mW_0, \vv_0$.

\paragraph{Bound on the training error:}
We start with a results that controls the training error and ensures that the coefficients of the neural network move little from its initialization.

\begin{theorem}
\label{thm:maintraining}
Pick an error parameter $\xi \in (0,1)$ and any failure probability $\delta \in (0,1)$, and choose $\nu,\omega$ so that they satisfy
$\nu \omega \leq  \xi / \sqrt{32\log(2n/\delta)}$. 
Suppose that the network is sufficiently overparameterized, i.e., 
\begin{align}
\label{eq:condk}
k \geq \Omega \left( 
\frac{n^{10} (\nu+\omega)^9}{\alpha^{11} \min(\nu,\omega) \xi^4}
\right).
\end{align}
\begin{enumerate}
\item[i)] 
Then, with probability at least $1-\delta$, the mean squared loss after $\iter$ iterations of gradient descent obeys
\begin{align}
\label{eq:boudtrainingerror}
\sqrt{\sum_{i=1}^n(y_i - f_{\mW_\iter,\vv_\iter}(\vx_i))^2}
\leq
\sqrt{ \sum_{i=1}^n \left(1 - \stepsize \sigma^2_i\right)^{2\iter} \innerprod{\vu_i}{\vy}^2
}
+ \xi \norm[2]{\vy}.
\end{align}
\item[ii)]
Moreover, the coefficients overall deviate little from its initialization, i.e.,
\begin{align}
\label{eq:movefrominit}
\sqrt{ \norm[F]{\mW_\iter - \mW_0}^2 + \norm[2]{\vv_\iter - \vv_0}^2 }
\leq
\underbrace{
\sqrt{
\sum_{i=1}^n
\left(
\innerprod{\vu_i}{\vy}
\frac{1 - (1 - \stepsize \sigma^2_i)^\iter}{\sigma_i}
\right)^2
}
+
\frac{\xi}{\alpha} \sqrt{n}
}_{Q \defeq }.
\end{align}
Here, $\norm[F]{\cdot}$ denotes the Frobenius norm.
In addition each of the coefficients changes only little, i.e., for all iterations $\iter$
\begin{align}
\label{eq:littlewtr}
\norm[2]{ \vw_{\iter,r} - \vw_{0,r} }
&\leq
(\nu + \frac{4}{\alpha} \sqrt{n}) \frac{n}{\sqrt{k}}
\frac{2}{\alpha^2}, \\
|v_{\iter,r} - v_{0,r}|
&\leq
\left(
O( \omega \sqrt{\log(n k / \delta)} )
+
\frac{4}{\alpha}
\sqrt{n}  \right)
\frac{n}{\sqrt{k}}
\frac{2}{\alpha^2}.
\label{eq:littlevtr}
\end{align}
Here, $\vw_{\iter,r}$ is the $r$-th row of $\mW_\iter$, and $\vv_{\iter,r}$ is the $r$-th entry of $\vv_\iter$. 
\end{enumerate}
\end{theorem}

\paragraph{Bound on the empirical risk:}
Because we train with respect to the $\ell_2$-loss but define the risk with respect to the (generic Lipschitz) loss $\loss$, the empirical risk and training loss are not the same. 
Nevertheless, we can upper bound the empirical risk computed over the training set at iteration $\iter$ with the training loss at iteration $\iter$:
\begin{align*}
\emprisk(f_{\mW_\iter, \vv_\iter})
&= 
\frac{1}{n} \sum_{i=1}^n \loss( f_{\mW_\iter, \vv_\iter} (\vx_i), y_i  ) \\
&\mystackrel{(i)}{\leq}
\frac{1}{n} \sum_{i=1}^n | f_{\mW_\iter, \vv_\iter} (\vx_i) - y_i  | \\
&\leq
\sqrt{
\frac{1}{n}
(f_{\mW_\iter, \vv_\iter}(\vx_i) - y_i)^2
} \\
&\mystackrel{(ii)}{\leq}
\sqrt{
\frac{1}{n}
\sum_{i=1}^n \innerprod{\vu_i}{\vy}^2 (1 - \eta \sigma_i^2)^{2\iter}
}
+ \xi,
\end{align*}
where 
(i) follows from
$\loss(z,y) = \loss(z,y) - \loss(y,y) \leq |z- y|$ 
because the loss is $1$-Lipschitz. 
Equation~(ii) is the most interesting one, and follows from Theorem~\ref{thm:maintraining}, equation~\eqref{eq:boudtrainingerror}, and holds with probability at least $1-\delta$. This bound is proven by showing that, provided the network is sufficiently wide, the training loss behaves as gradient descent applied to a linear least-squares problem with dynamics governed by the gram matrix $\mSigma$.


\paragraph{Bound on the generalization error:}
Next, we bound the generalization error $\risk(f) - \emprisk(f)$ by bounding the Rademacher complexity of the functions that gradient descent can reach with $\iter$ gradient descent iterations.

Let $\mc F$ be a class of functions $f \colon \reals^d \to \reals$. Let $\epsilon_1,\ldots,\epsilon_n$ be iid Rademacher random variables, i.e., random variables that are chosen uniformly from $\{-1,1\}$. Given the dataset $\setD$, define the \emph{empirical Rademacher complexity} of the function class $\setF$ as 
\begin{align*}
\mc R_{\setD}(\setF)
=
\frac{1}{n}
\EX[\vepsilon]{ \sup_{f \in \setF} \sum_{i=1}^n \epsilon_i f(\vx_i) }.
\end{align*}
Here, $\setD =\{(\vx_1,y_1),\ldots, (\vx_n,y_n)\}$ is the training set, consisting of $n$ points drawn iid from the example generating distribution. 
By a standard result from statistical learning theory, a bound on the Radermacher complexity directly gives a bound on the generalization error for each predictor in a class of predictors. 

\begin{theorem}[{ \cite[Thm.~3.1]{mohri_FoundationsMachineLearning_2012} }]
Suppose $\loss(\cdot,\cdot)$ is bounded in $[0,1]$ and $1$-Lipschitz in its first argument. 
With probability at least $1-\delta$ over the random dataset $\setD$ consisting of $n$ iid examples, we have that
\begin{align*}
\sup_{f \in \setF}
\risk(f) - \emprisk(f)
\leq
2 \mc R_{\setD}(\setF) 
+ 3 \sqrt{ \frac{\log(2/\delta)}{2n} }.
\end{align*}
\end{theorem}

We consider the class of neural networks with weights close to the random initialization $\mW_0,\vv_0$, defined as:
\begin{align}
\label{eq:defclassnetworks}
\setF_{Q,M}
=
\left\{
f_{\mW,\vv}
\colon  
\mW \in \setW, \vv \in \setV \right\}, 
\end{align}
with 
\begin{align*}
\setW 
&=
\left\{ \mW \colon \norm[F]{\mW - \mW_0} \leq  Q, \norm[2]{\vw_r - \vw_{0,r}} \leq \omega M, \text{ for all $r$}
\right\},\\
\setV 
&=
\left\{ \vv \colon \norm[2]{\vv - \vv_0} \leq Q, |v_r - v_{0,r}| \leq \nu M, \text{ for all $r$}
\right\}.
\end{align*}
The Rademacher complexity of this class of functions is controlled with the following result.

\begin{lemma}
\label{lem:lemrademacher}
Let $\mW_0$ be drawn from a Gaussian distribution with $\mc N(0,\omega^2)$ entries, and suppose the entries of $\vv_0$ are draw uniformly from $\{-\nu, \nu\}$. 
Assume the $(\vx_i,y_i)$ are drawn iid from some distribution with $\norm[2]{\vx_i} =1$ and $|y_i| \leq  1$. 
With probability at least $1-\delta$ over the random training set, provided that $\sqrt{\log(2n/\delta) / 2 k} \leq 1/2$, the empirical Rademacher complexity of $\setF_{Q,M}$ is, simultaneously for all $Q$, bounded by
\begin{align}
\setR_\setD(\setF_{Q,M})
\leq
\frac{Q}{\sqrt{n}} (\nu+\omega)
+
\nu \omega (5M^2 \sqrt{k} + 4M \sqrt{\log(2/\delta)/2}).
\end{align}
\end{lemma}

We set $M = O(\frac{\xi}{\alpha} k^{-1/4})$. With this choice,
the term on the right hand side above is bounded by
\[
\nu \omega (5M^2 \sqrt{k} + 4M \sqrt{\log(2/\delta)/2})
\leq
O(\xi/\alpha),
\]
where we used $\nu \omega \leq 1$ and $\frac{\sqrt{\log(2/\delta)/2}}{k^{1/4}} \leq 1$, by assumption~\eqref{eq:condkmainthm}. 
Note that by~\eqref{eq:littlewtr} and by~\eqref{eq:littlevtr} combined with the assumption~\eqref{eq:condkmainthm} we have that $\norm[2]{\vw_r - \vw_{0,r}} \leq \omega M$ and $|v_r - v_{0,r}| \leq \nu M$, as desired.

Let $Q_i = i $ for $i = 1, 2,\ldots$. Simultaneously for all $i$, by the lemma above, for this choice of $M$, the function class $\setF_{Q_i,M}$ has Rademacher complexity bounded by 
\begin{align}
\setR_\setD(\setF_{Q_i,M})
\leq
\frac{Q_i}{\sqrt{n}} (\nu+\omega)
+
O(\xi/\alpha).
\end{align}
We next choose the radius $Q$ as defined in~\eqref{eq:movefrominit}. 
Let $i^\ast$ be the smallest integer such that $Q \leq Q_{i^\ast}$, so that $Q_{i^\ast} \leq Q+1$. 
We have that $i^\ast \leq O(\sqrt{n/\alpha})$ and
\begin{align}
\setR_\setD(\setF_{Q_{i^\ast},M})
&\leq
\frac{(Q+1)}{\sqrt{n}} (\nu+\omega)
+
O(\xi/\alpha) \nonumber \\
&\leq
\sqrt{
\frac{1}{n}
\sum_{i=1}^n
\left(
\innerprod{\vu_i}{\vy}
\frac{1 - (1 - \stepsize \sigma^2_i)^\iter}{\sigma_i}
\right)^2
}
+
\frac{1}{\sqrt{n}} 
+
O(\xi/\alpha),
\end{align}
by the assumption of the theorem on $k$ being sufficiently large, and by $\nu + \omega \leq 1$. 
Next, from a union bound over the finite set of integers $i= 1,\ldots, i^\ast$, we obtain
\begin{align}
\max_{i= 1,\ldots, i^\ast}
\sup_{f \in \setF_{Q_i,M}}
\risk(f) - \emprisk(f)
\leq
\sqrt{
\frac{1}{n}
\sum_{i=1}^n
\left(
\innerprod{\vu_i}{\vy}
\frac{1 - (1 - \stepsize \sigma^2_i)^\iter}{\sigma_i}
\right)^2
}
+
\frac{1}{\sqrt{n}}
+
O(\xi/\alpha),
\end{align}
as desired.

%

%
\paragraph{Final bound on the risk:}
Combining the bound on the training with the generalization bound yields the upper bound~\eqref{eq:finalriskbound} on the risk of the network trained for $\iter$ iterations of gradient descent.

The remainder of the proof is devoted to proving Theorem~\ref{thm:maintraining} and Lemma~\ref{lem:lemrademacher}.  

\subsection{Preliminaries}

We start with introducing some useful notation.
First note that the prediction of the neural network for the $n$ training data points as a function of the parameters are
\begin{align}
\label{eq:predictions}
\vf(\mW,\vv) 
= 
\frac{1}{\sqrt{k}}
\begin{bmatrix}
\relu(\transp{\vx}_1 \mW)\vv \\
\vdots \\
\relu(\transp{\vx}_n \mW)\vv
\end{bmatrix}
=
\frac{1}{\sqrt{k}}
\relu(\mX \mW) \vv,
\end{align}
where $\mX^{n\times d}$ is the feature matrix and $\mW \in \reals^{d\times k}$ and $\vv \in \reals^k$ are the trainable weights of the network.
The transposed Jacobian of the function $\vf$ is given by
\begin{align}
\label{eq:jacobian}
\transp{\Jac}(\mW,\vv) 
= 
\begin{bmatrix}
\transp{\Jac}_1(\mW,\vv) \\
\transp{\Jac}_2(\mW)
\end{bmatrix}
\in \reals^{dk + k \times n},
\end{align}
where we defined the Jacobians corresponding to the weights of the first layer, $\mW$, and the second layer, $\vv$, respectively as
\begin{align*}
\quad
\transp{\Jac}_1(\mW,\vv)
=
\frac{1}{\sqrt{k}}
\begin{bmatrix}
v_1 \transp{\mX} \diag(\relu'(\mX\vw_1)) \\
\vdots \\
v_\iter \transp{\mX} \diag(\relu'(\mX\vw_\iter))
\end{bmatrix}
\in \reals^{dk \times n}
,
\quad
\transp{\Jac}_2(\mW)
=
\frac{1}{\sqrt{k}}
\transp{ \relu(\mX \mW) } \in \reals^{k\times n}.
\end{align*}
Here, $\relu'(x) = \ind{x \geq 0}$ is the derivative of the relu activation function, which is the step function.
Our results depend on the singular values and vectors of the expected Jacobian at initialization:
\begin{align*}
\EX{ \Jac(\mW_0,\vv_0) \transp{\Jac}(\mW_0,\vv_0) }
&=
\nu^2
\sum_{\ell=1}^k
\EX{ \relu'(\mX \vw_{0,\ell}) \transp{ \relu'(\mX \vw_{0,\ell}) }  } \odot \mX \transp{\mX} \\
&+
\frac{1}{k}
\EX{\relu(\mX \mW_0) \relu(\mX \mW_0)^T},
\end{align*}
where $\odot$ is the Hadamard product, and where we used that the entries of  $\vv_0$ are choosen iid uniformly from $\{-\nu ,  \nu \}$.
Expectation is over the weights $\mW_0$ at initialization, which are iid $\mc N(0, \omega^2 )$. 
This yields
\begin{align}
\left[\EX{ \Jac(\mW_0,\vv_0) \transp{\Jac}(\mW_0,\vv_0) }\right]_{ij}
&=
\nu^2
K_1(\vx_i,\vx_j)
+
\omega^2
K_2(\vx_i,\vx_j),
\end{align}
where $K_1$ and $K_2$ are two kernels associated with the first and second layers of the network and are given by 
\begin{align*}
K_1(\vx_i, \vx_j)
&=
\left[
\EX{ \relu'(\mX \vw_\ell) \transp{ \relu'(\mX \vw_\ell) } }
\right]_{ij} \\
&=
\frac{1}{2}
\left(
1 - \cos^{-1} \left( \rho_{ij} \right) / \pi
\right)
\innerprod{\vx_i}{\vx_j}
\end{align*}
with $\rho_{ij} = \frac{\innerprod{\vx_i}{\vx_j}}{ \norm[2]{\vx_i} \norm[2]{\vx_j} }$ and by 
\begin{align*}
 K_2(\vx_i, \vx_j)
&=
\frac{1}{\omega^2}
\frac{1}{k}
\left[ \EX{\relu(\mX \mW) \relu(\mX \mW)^T} \right]_{ij} \\
&=
\frac{1}{\omega^2} \left[ 
\EX{\relu(\mX \vw) \relu(\mX \vw)^T} \right]_{ij} \\
&=
\frac{1}{2}
\left(
\sqrt{1 - \rho_{ij}^2 }/\pi
+
(1 - \cos^{-1}(\rho_{ij})/\pi ) \rho_{ij} 
\right) \norm[2]{\vx_i} \norm[2]{\vx_j}.
\end{align*}
For both of those expressions, we used the calculations from~\cite[Sec.~4.2]{daniely_DeeperUnderstandingNeural_2016} for the final expressions of the kernels.
Also note that, by assumption $\norm[2]{\vx_i} = 1$.


\subsection{Proof of Theorem~\ref{thm:maintraining} (bound on the training error)}

In this subsection, we prove Theorem~\ref{thm:maintraining}. 


\subsubsection{The dynamics of linear and nonlinear least-squares}\label{metathms}

\newcommand\N{N}
\newcommand\losslin{\mc L_{\mathrm{lin}}}
\newcommand\flin{f_{\mathrm{lin}}}

Theorem~\ref{thm:maintraining} relies on approximating the trajectory of gradient descent applied to the training loss with an associated linear model that approximates the non-linear neural network in the highly-overparameterized regime.
This strategy has been used in a number of recent publications~\citep{arora_FineGrainedAnalysisOptimization_2019,du_GradientDescentProvably_2018b,
oymak_ModerateOverparameterizationGlobal_2020,
oymak_GeneralizationGuaranteesNeural_2019,heckel_DenoisingRegularizationExploiting_2020}; 
in order to avoid repetition, we rely on a statement~\cite[Theorem 4]{heckel_CompressiveSensingUntrained_2020}, which bounds the error between the true trajectory of gradient descent and the trajectory of an associated linear problem.

Let $\vf\colon \reals^\N \rightarrow \reals^n$ be a non-linear function with parameters $\vtheta \in \reals^\N$, and consider the non-linear least squares problem
\[
\mc L(\vtheta) = \frac{1}{2} \norm[2]{ \vf(\vtheta) - \vy}^2.
\]
The gradient descent iterations starting from an initial point $\vtheta_0$ are given by
\begin{align}
\label{iterupdates}
\vtheta_{\iter + 1} = \vtheta_\iter - \stepsize \nabla \mc L(\vtheta_\iter)
\quad \text{where}\quad
\nabla \mc L(\vtheta)
= \transp{\Jac}(\vtheta) ( \vf(\vtheta) - \vy),
\end{align}
where $\Jac(\vtheta)\in\reals^{n\times N}$ is the Jacobian of $\vf$ at $\vtheta$ (i.e., 
$[\Jac(\vtheta)]_{i,j} = \frac{\partial \vf_i(\vtheta)}{ \partial \vtheta_j }$). 
The associated linearized least-squares problem is defined as
\begin{align}
\label{linprob}
\losslin(\vtheta)
=
\frac{1}{2}
\norm[2]{ \vf(\vtheta_0) + \mJ (\vtheta - \vtheta_0) - \vy}^2.
\end{align}
Here, $\mJ\in\reals^{n\times \N}$, refered to as the reference Jacobian, is a fixed matrix independent of the parameter $\vtheta$ that approximates the Jacobian mapping at initialization, $\Jac(\vtheta_0)$. 
Starting from the same initial point $\vtheta_0$, the gradient descent updates of the linearized problem are
\begin{align}
\label{iterupdateslin}
\tilde{\vtheta}_{\iter + 1}
&= 
\tilde{\vtheta}_\iter - \stepsize \transp{\mJ} \left( \vf(\vtheta_0) + \mJ (\tilde{\vtheta}_\iter - \vtheta_0) - \vy\right).
\end{align}
To show that the non-linear updates~\eqref{iterupdates} are close to the  linearized iterates~\eqref{iterupdateslin}, we make the following assumptions:

\begin{subequations}
\begin{enumerate}
\item[i)] We assume that the singular values of the reference Jacobian obey for some $\alpha,\beta$
\begin{align}
\label{eq:Jacbounded0}
\sqrt{2} \alpha \le \sigma_{n} \le \sigma_{1} \le \beta.
\end{align}
Furthermore, we assume that the norm of the Jacobian associated with the nonlinear model $\vf$ is bounded in a radius $R$ around the random initialization
\begin{align}
\label{eq:Jacbounded}
\norm{\Jac(\vtheta)} \leq \beta\quad\text{for all}\quad \vtheta\in\ball_R(\vtheta_0).
\end{align}
Here, $\ball_R(\vtheta_0) \defeq \{\vtheta \colon \norm{\vtheta - \vtheta_0} \leq R\}$ is the ball with radius $R$ around $\vtheta_0$.
\item[ii)] We assume the reference Jacobian and the Jacobian of the nonlinearity at initialization $\Jac(\vtheta_0)$ are $\epsilon_0$-close:
\begin{align}
\label{eq:assJacJclose}
\norm{\Jac(\vtheta_0) - \mJ} \leq \epsilon_0.
\end{align}
\item[iii)]
We assume that within a radius $R$ around the initialization, the Jacobian varies by no more than $\epsilon$:
\begin{align}
\label{eq:Jacclose}
\norm{\Jac(\vtheta) - \Jac(\vtheta_0)}
\leq
\frac{\epsilon}{2},
\quad 
\text{for all}
\quad
\vtheta \in \ball_R(\vtheta_0).
\end{align}
\end{enumerate}
\end{subequations}
Under these assumptions the difference between the non-linear residual 
\begin{align*}
\vr_\iter \defeq f(\vtheta_\iter) - \vy
\end{align*}
and the linear residual
\begin{align*}
\tilde \vr_\iter \defeq
f(\vtheta_0) + \mJ (\tilde{\vtheta}_\iter - \vtheta_0) - \vy
\end{align*}
are close throughout the entire run of gradient descent. 

\begin{theorem} [{\cite[Theorem 4]{heckel_CompressiveSensingUntrained_2020}}, Closeness of linear and nonlinear least-squares problems]
\label{thm:maintechnical}
Assume the Jacobian $\Jac(\vtheta)\in\reals^{n\times \N}$ associated with the function $\vf(\vtheta)$ obeys Assumptions \eqref{eq:Jacbounded0}, \eqref{eq:Jacbounded}, \eqref{eq:assJacJclose}, and \eqref{eq:Jacclose} around an initial point $\vtheta_0\in\reals^\N$ with respect to a reference Jacobian $\mJ\in\reals^{n\times \N}$ and with parameters $\alpha, \beta, \epsilon_0, \epsilon$, obeying 
$2\beta (\epsilon_0 + \epsilon) \leq \alpha^2$, and $R$. Furthermore, assume the radius $R$ is given by
\begin{align}
\label{eq:defR}
R
\defeq
2\norm[2]{ \mJ^\dagger\vr_0}
+ 5\frac{\beta^2}{\alpha^4}(\epsilon_0 + \epsilon)   \norm[2]{\vr_0}.
\end{align}
Here, $\pinv{\mJ}$ is the pseudo-inverse of $\mJ$. 
We run gradient descent with stepsize $\stepsize \leq \frac{1}{\beta^2}$ on the linear and non-linear least squares problem, starting from the same initialization $\vtheta_0$. 
Then, for all iterations $\iter$,
\begin{enumerate}
\item[i)] the non-linear residual converges geometrically
\begin{align}
\label{resnl}
\norm[2]{\vr_{\iter}}\le \left(1-\eta\alpha^2\right)^t \norm[2]{\vr_0},
\end{align}
\item[ii)]  
the residuals of the original and the linearized problems are close
\begin{align}
\norm[2]{\vr_\iter - \tilde \vr_\iter} 
%
&\leq
\frac{2\beta (\epsilon_0 + \epsilon)}{e(\ln{2}) \alpha^2}
\norm[2]{\vr_0},\label{eq:propvrkclose2}
\end{align}
\item[iii)]  the parameters of the original and the linearized problems are close
\begin{align}
\norm[2]{\vtheta_\iter - \tilde{\vtheta}_\iter} 
&\leq 
2.5 \frac{\beta^2}{\alpha^4} (\epsilon_0 + \epsilon)\norm[2]{\vr_0},
\label{eq:propthetakclose}
\end{align}
\item[iv)]
and the parameters are not far from the initialization 
\begin{align}
\label{eq:indhypinRBall}
\norm[2]{\vtheta_\iter - \vtheta_0} \leq \frac{R}{2}.
\end{align}
\end{enumerate}
\end{theorem}
Theorem~\ref{thm:maintechnical} above formalizes that in a (small) radius around the initialization, the non-linear problem behaves very similar to its associated linear problem.
As a consequence, to characterize the dynamics of the nonlinear problem, it suffices to characterize the dynamics of the linearized problem. This is the subject of our next theorem, which is a standard result on the gradient iterations of a least squares problem, see for example~\citep[Thm.~5]{heckel_DenoisingRegularizationExploiting_2020} for the proof.
%
\begin{theorem}[{E.g. Theorem~5 in \cite{heckel_DenoisingRegularizationExploiting_2020}}]
\label{lem:rk}
Consider a linear least squares problem~\eqref{linprob} and let $\mJ 
=\sum_{i=1}^n\sigma_i \vu_i \vv_i^T$
be the singular value decomposition of the matrix $\mJ$.  Then the linear residual $\tilde \vr_\iter$ after $\iter$ iterations of gradient descent with updates~\eqref{iterupdateslin} is
\begin{align}
\label{eq:linresformula}
\tilde \vr_\iter 
= 
\sum_{i=1}^n \left(1 - \stepsize \sigma^2_i\right)^\iter \vu_i \innerprod{\vu_i}{\vr_0}.
\end{align}
Moreover, using a step size satisfying $\stepsize \leq \frac{1}{\sigma_1^2}$, the linearized iterates \eqref{iterupdateslin} obey
\begin{align}
\label{eq:lindifftheta}
\norm[2]{\tilde{\vtheta}_\iter - \vtheta_0}^2
=
\sum_{i=1}^n
\left(
\innerprod{\vu_i}{\vr_0}
\frac{1 - (1 - \stepsize \sigma^2_i)^\iter}{\sigma_i}
\right)^2.
\end{align}
\end{theorem}


\subsubsection{Proving Theorem~\ref{thm:maintraining} by applying Theorem~\ref{thm:maintechnical}}

We are now ready to prove Theorem~\ref{thm:maintraining}. 
We apply Theorem~\ref{thm:maintechnical} to the predictions of the network given by $\vf(\mW,\vv)$ defined in~\eqref{eq:predictions} with parameter $\vtheta = (\mW,\vv)$. 
As reference Jacobian we choose a matrix $\mJ \in \reals^{n \times dk + k}$ that satisfies $\mJ \transp{\mJ} = \EX{ \Jac(\mW_0,\vv_0) \transp{\Jac}(\mW_0,\vv_0) }$ (where expectation is over the random initialization $(\mW_0,\vv_0)$), and at the same time is very close to the Jacobian of $\vf$ at initialization, i.e., to $\Jac(\mW_0,\vv_0)$. 
Towards this goal, we apply Theorem~\ref{thm:maintechnical} with the following choices of parameters:
\begin{align}
\alpha = \sigma_{\min}(\mSigma) / \sqrt{2},
\quad
\beta = 10 \sqrt{n} (\omega + \nu),
\quad
\epsilon = \frac{1}{16} \xi \frac{\alpha^3}{\beta^2},
\quad
\epsilon_0 = 2 \sqrt{(\omega^2 + \nu^2) \frac{3n}{\sqrt{k}} \log(kn/\delta)  }.
\end{align}
Note that assumption~\eqref{eq:condk} guarantees that $\epsilon_0 \leq \epsilon$, a fact we used later.

We now verify that the conditions of Theorem~\ref{thm:maintraining} are satisfied for this choice of parameters with probability at least $1-\delta$.
Specifically we show that each of the conditions holds with probability at least $1-\delta$. By a union bound, the success probability is then at least $1-\Omega(\delta)$, and by rescaling $\delta$ by a constant, the conditions are satisfied with probability at least $1-\delta$. 


\paragraph{Bound on residual:}
We need a bound on the network outputs at initialization as well as on the initial residual to verify the conditions of the theorem.
We start with the former:
\begin{align}
\norm[2]{\vf(\mW_0,\vv_0)}
&=
\frac{1}{\sqrt{k}} \norm[2]{\relu(\mX \mW_0)\vv_0} \nonumber \\
&\leq
\nu \omega
\sqrt{8\log(2n/\delta)} \norm[F]{\mX} \nonumber \\
&= \nu \omega
\sqrt{8\log(2n/\delta)} \sqrt{n},
\label{eq:boundinitial}
\end{align}
where the inequality holds with probability at least $1-\delta$, by Gaussian concentration (see Lemma~6 in~\cite{heckel_DenoisingRegularizationExploiting_2020} and recall that $\mW_0$ has iid $\mc N(0,\omega^2)$ entries).
Moreover, the last equality follows from $\norm[2]{\vx_i} = 1$. 

It follows that, with probability at least $1-\delta$, the initial residual is bounded by
\begin{align}
\norm[2]{\vr_0}
&=
\norm[2]{ \frac{1}{\sqrt{k}} \relu(\mX \mW_0)\vv_0 - \vy} \nonumber \\
&\leq
\nu \omega 
\sqrt{8\log(2n/\delta)} \sqrt{n}
+
\sqrt{n} \nonumber \\
&\leq 2 \sqrt{n}
\label{eq:boundresidual}
\end{align}
where the first inequality holds by the triangle inequality and using the assumption $|y_i|\leq 1$,
 and the second inequality by $\nu \omega 
\sqrt{8\log(2n/\delta)} \leq 1$, again by assumption.

\paragraph{Radius in the theorem:}
In order to verify the condition of the theorem, we need to control the radius in the theorem, which we do next. 
With our assumptions and the choices of parameters above, the radius in the theorem, defined in equation~\eqref{eq:defR}, obeys
\begin{align}
R 
&=
2 \norm[2]{ \mJ^\dagger\vr_0}
+ 5\frac{\beta^2}{\alpha^4}(\epsilon_0 + \epsilon)   \norm[2]{\vr_0} \nonumber\\
&\mystackrel{(i)}{\le}
\left(
\frac{\sqrt{2}}{\alpha} 
+
5\frac{\beta^2}{\alpha^4}(\epsilon_0 + \epsilon)
\right) \norm[2]{\vr_0} \nonumber \\
&\mystackrel{(ii)}{\le}
\left(
\frac{\sqrt{2}}{\alpha} 
+
\frac{5}{16\alpha}
\right) \norm[2]{\vr_0} \nonumber \\
&\mystackrel{(iii)}{\leq}
\frac{4}{\alpha} 
\sqrt{n} \nonumber \\
&\mystackrel{(iv)}{\leq}
\min(\omega,\nu)
\underbrace{
\sqrt{k} \left(\frac{ \frac{1}{16} \xi \frac{\alpha^3}{\beta^2} }{\norm{\mX} (\nu + 3\omega) }\right)^3
}_{\tilde{R}}.
\label{eq:boundradius}
\end{align}
Here,
(i) follows from the fact that $\norm[2]{\mJ^\dagger\vr_0}\le \frac{1}{\sqrt{2}\alpha}\norm[2]{\vr_0}$, 
(ii) from $\epsilon_0 + \epsilon \leq 2 \epsilon = \frac{1}{8} \xi \frac{\alpha^3}{\beta^2}$ (by definition of $\epsilon_0$ and $\epsilon$),
and 
(iii) from the bound on the residual~\eqref{eq:boundresidual}. 
For (iv) we used assumption~\eqref{eq:condk} in the theorem.


\paragraph{Verifying Assumptions~\eqref{eq:Jacbounded0} and \eqref{eq:Jacbounded}:} 
By definition $\mJ\mJ^T= \mSigma$, thus the lower bound in assumption~\eqref{eq:Jacbounded0} holds by the definition of $\alpha$ as $\sigma_{n} (\mSigma) \geq \sqrt{2}\alpha$. 
Regarding the upper bound of \eqref{eq:Jacbounded0}, note that
\begin{align*}
\norm{\mJ}^2
&\leq
\nu^2 \norm[F]{\mK_1}
+
\omega^2 \norm[F]{\mK_2} \\
&\leq
\nu^2 n
+
\omega^2 n,
\end{align*}
where $\mK_1 \in \reals^{n\times n}$ and $\mK_2 \in \reals^{n\times n}$ are the kernel matrix with entries $K_1(\vx_i,\vx_j)$ and $K_2(\vx_i,\vx_j)$. 
It follows that 
$\norm{\mJ} \leq 2(\omega + \nu) \sqrt{n} \leq \beta$, as desired.
This concludes the verification of \eqref{eq:Jacbounded0}. 
 
To verify assumption~\eqref{eq:Jacbounded} note that
\begin{align*}
\norm{\Jac(\mW,\vv)}
&\leq
\norm{\Jac_1(\mW)}
+
\norm{\Jac_2(\mW)} \\
&\leq
\frac{1}{\sqrt{k}}
 ( \norm{\mX} \norm[2]{\vv} + \norm[F]{\mX \mW} ) \\
&\leq
\frac{1}{\sqrt{k}}
 \left(
 \norm{\mX}( \norm[2]{\vv_0}  + \norm[2]{\vv - \vv_0}  + \norm[F]{\mW - \mW_0} )  + \norm[F]{\mX \mW_0} \right) \\
&\leq
 \sqrt{n} 10 (\omega + \nu) = \beta.
\end{align*}
For the last inequality, we used that 
$\norm[2]{\vv_0} = \nu \sqrt{k}$, 
and that $\norm[2]{\vv - \vv_0} + \norm[F]{\mW - \mW_0} \leq  R \leq \sqrt{k} \min(\omega,\nu)$, by the bound on the radius in~\eqref{eq:boundradius},
 and finally that 
 $\norm[F]{\mX \mW_0} \leq \omega 6 \sqrt{k}$ with probability at least $1-\delta$ provided that $k \geq \log(n/\delta)$, which holds by assumption. 
 For this inequality we used that $\transp{\vx}_i \mW_0$ is a Gaussian vector with iid $\mc N(0,\omega^2)$ entries. 
It follows that assumption~\eqref{eq:Jacbounded} holds with probability at least $1-\delta$, as desired.


\paragraph{Verifying Assumption~\eqref{eq:assJacJclose}:}
We start with stating a concentration lemma from~\cite{heckel_DenoisingRegularizationExploiting_2020}.

\begin{lemma}[Concentration lemma {\citep[Lemma~3]{heckel_DenoisingRegularizationExploiting_2020}}]
\label{lem:ConcentrationLemma}
Consider the partial Jacobian $\Jac_1(\mW)$, and let $\mW\in\reals^{n\times k}$ be generated at random with i.i.d.~$\mathcal{N}(0,\omega^2)$ entries, and suppose the $v_\ell$ are drawn from a distribution with $|v_\ell| \leq \nu$.
Then, with probability at least $1-\delta$,
\begin{align*}
\norm{ \Jac_1(\mW,\vv)\Jac^T_1(\mW,\vv) - \EX{ \Jac_1(\mW,\vv)\Jac^T_1(\mW,\vv) } }
\leq
\frac{\nu^2}{\sqrt{k}}
\norm{\mX}^2 \sqrt{ \log\left(2n/\delta\right)} .
\end{align*}
\end{lemma}

\begin{lemma}
\label{lem:ConcentrationLemma2}
Let $\Jac_2(\mW) = \frac{1}{\sqrt{k}} \relu(\mX \mW)$, with $\mW$
generated at random with i.i.d.~$\mathcal{N}(0,\omega^2)$ entries. 
With probability at least $1-\delta$, 
\begin{align}
\label{eq:fromconcentration2}
\norm{ \Jac_2(\mW)\Jac^T_2(\mW) - \EX{ \Jac_2(\mW)\Jac^T_2(\mW) } }
\leq
3\frac{\omega^2}{\sqrt{k}} \norm{\mX}^2 \log(kn / \delta).
\end{align}
\end{lemma}

Combining the statements of the two lemmas, it follows that, with probability at least $1-2\delta$,
\begin{align}
\norm{ \Jac(\mW,\vv)\Jac^T(\mW,\vv) - \EX{ \Jac(\mW,\vv)\Jac^T(\mW,\vv) } }
&\leq
(\omega^2 + \nu^2) 3\frac{1}{\sqrt{k}} \norm{\mX}^2 \log(kn / \delta)  \nonumber \\
&\leq
(\omega^2 + \nu^2) 3\frac{1}{\sqrt{k}} n
\log(kn / \delta).
\label{eq:fromconcentration}
\end{align}
To show that \eqref{eq:fromconcentration} implies the condition in~\eqref{eq:assJacJclose}, we use the following lemma.
\begin{lemma}[{\citep[Lem.~6.4]{oymak_GeneralizationGuaranteesNeural_2019}}]
\label{eq:lemmaC1impC2}
Let  $\mJ_0 \in \reals^{n\times \N}$, $\N \geq n$ and let $\mSigma$ be $n \times n$ psd matrix obeying $\norm{ \mJ_0 \transp{\mJ}_0 - \mSigma } \leq \tilde \epsilon^2$, for a scalar $\tilde \epsilon \geq 0$. Then there exists a matrix $\mJ \in \reals^{n \times \N}$ obeying $\mSigma = \mJ \transp{\mJ}$ such that
\[
\norm{\mJ - \mJ_0} \leq 2 \tilde \epsilon.
\]
\end{lemma}

From Lemma~\ref{eq:lemmaC1impC2} combined with equation~\eqref{eq:fromconcentration}, there exists a matrix $\mJ \in \reals^{n\times N}$ that obeys 
\begin{align*}
\norm{ \mJ - \Jac(\mW_0,\vv_0)} 
\leq 
\epsilon_0,
\quad \epsilon_0 = 2 \sqrt{(\omega^2 + \nu^2) \frac{3n}{\sqrt{k}} \log(kn/\delta)  }.
\end{align*}

This part of the proof also specifies our choice of the matrix $\mJ$ as a matrix that is $\epsilon_0$ close to the Jacobian at initialization, $\Jac(\mW_0)$, and that exists by Lemma~\ref{eq:lemmaC1impC2} above.

\paragraph{Verifying Assumption~\eqref{eq:Jacclose}:}
We control the perturbation around the random initialization. 

\begin{lemma}
\label{eq:lemchangeJac}
Let $\mW_0$ have iid $\mc N(0,\omega^2)$ entries and let $\vv_0$ have (arbitrary) entries in $\{-\nu , +\nu \}$ entries. Then for all $\mW$ and $\vv$ obeying, for some $\tilde R \leq \frac{1}{2} \sqrt{k}$, 
\begin{align*}
\norm[F]{\mW - \mW_0} \leq \omega \tilde R,
\quad
\norm[2]{\vv - \vv_0} \leq \nu \tilde R,
\end{align*}
the Jacobian in~\eqref{eq:jacobian} obeys
\begin{align*}
\norm{\Jac(\mW,\vv) - \Jac(\mW_0,\vv_0)}
\leq
\norm{\mX}
\frac{1}{\sqrt{k}}
\left(
\omega \tilde R
+
\nu \tilde R
+
\nu \sqrt{2} (2\iter \tilde R)^{1/3}
\right),
\end{align*}
with probability at least $1- n e^{- \frac{1}{2} \tilde R^{4/3} k^{7/3} }$. 
\end{lemma}

Recall the definition $\tilde R = \sqrt{k} \left(\frac{\epsilon}{\norm{\mX} (\nu + 3\omega) }\right)^3$ from~\eqref{eq:boundradius}. 
From $\tilde R \leq (k R)^{1/3}$ for $\tilde R \leq \sqrt{k}$, the bound provided by lemma~\ref{eq:lemchangeJac} guarantees that
\begin{align*}
\norm{\Jac(\mW,\vv) - \Jac(\mW_0,\vv_0)}
&\leq
\norm{\mX}
\frac{\omega + 3\nu}{\sqrt{k}}
(k \tilde R)^{1/3} \\
&= \epsilon = \frac{1}{16} \xi \frac{\alpha^3}{\beta^2},
\end{align*}
where the second inequality follows by choosing
 $\tilde R = \sqrt{k} \left(\frac{\epsilon}{\norm{\mX} (\nu + 3\omega) }\right)^3$.
This holds with probability at least
\begin{align*}
1-n e^{-\frac{1}{2} \tilde R k^{7/3}}
=
1-n e^{- 2^{-17} \xi^4 \frac{\alpha^8}{\beta^8} k^{3}}
\mystackrel{(i)}{\ge} 1-\delta,
\end{align*}
where in (i) we used~\eqref{eq:condk}. 
Therefore, Assumption~\eqref{eq:Jacclose} holds with high probability by our choice of $\epsilon = \frac{1}{16} \xi \frac{\alpha^3}{\beta^2}$.

\paragraph{Concluding the proof of Theorem~\ref{thm:maintraining}:}
By the previous paragraphs, the assumptions of Theorem~\ref{thm:maintraining} are satisfied with probability at least $1 - O(\delta)$. Therefore we can bound bound the training error and the deviation of the coefficients from the initialization as follows.

\paragraph{Training error:}
We bound the training error in~\eqref{eq:condk}. 
The training error at iteration $\iter$ is bounded by
\begin{align*}
\norm[2]{ \vf(\mW_\iter, \vv_\iter) - \vy }
&\leq
\norm[2]{ \tilde \vr_\iter }
+
\norm[2]{ \tilde \vr_\iter - \vr_\iter} \\
&\mystackrel{(i)}{\leq}
\sqrt{ \sum_{i=1}^n \left(1 - \stepsize \sigma^2_i\right)^{2\iter} \innerprod{\vu_i}{\vr_0}^2
}
+
\frac{2\beta (\epsilon_0 + \epsilon)}{e(\ln{2}) \alpha^2}
\norm[2]{\vr_0} \\
&\mystackrel{(ii)}{\leq}
\sqrt{ \sum_{i=1}^n \left(1 - \stepsize \sigma^2_i\right)^{2\iter} \innerprod{\vu_i}{\vy}^2
}
+
\norm[2]{\vf(\mW_0, \vw_0)} + \frac{2\beta (\epsilon_0 + \epsilon)}{e(\ln{2}) \alpha^2}
\norm[2]{\vr_0} \\
&\mystackrel{(iii)}{\leq}
\sqrt{ \sum_{i=1}^n \left(1 - \stepsize \sigma^2_i\right)^{2\iter} \innerprod{\vu_i}{\vy}^2
}
+ \xi \norm[2]{\vy},
\end{align*}
where inequality (i) follows from bounding the linear residual $\norm[2]{ \tilde \vr_\iter }$ with theorem~\ref{lem:rk}, as well as bounding the distance between the linear residual and the non-linear one with~\eqref{eq:propvrkclose2}. 
Inequality (ii) follows from $\vr_0 = \vf(\mW_0, \vv_0) - \vy$,
and finally (iii) follows from 
$
\norm[2]{\vf(\mW_0,\vv_0)}
\leq
\nu \omega
\sqrt{8\log(2n/\delta)} \norm[2]{\vy}
\leq
\frac{\xi}{2} \norm[2]{\vy}
$,
by \eqref{eq:boundinitial}, and 
$\frac{\beta}{\alpha^2} (\epsilon_0 + \epsilon)\leq
\frac{\beta}{\alpha^2} 2 \epsilon
=  \frac{1}{8} \xi \frac{\alpha}{\beta}
\leq
 \xi
$.

\paragraph{Distance from initialization:}
We next bound the distance from the initialization, i.e., we establish~\eqref{eq:movefrominit}.  Combining equation~\eqref{eq:lindifftheta} in theorem~\ref{lem:rk}
with equation~\eqref{eq:propvrkclose2} in theorem~\ref{thm:maintechnical}, we obtain
\begin{align*}
\sqrt{ \norm[F]{\mW_\iter - \mW_0}^2 + \norm[2]{\vv_\iter - \vv_0}^2 }
&\leq
\sqrt{
\sum_{i=1}^n
\left(
\innerprod{\vu_i}{\vr_0}
\frac{1 - (1 - \stepsize \sigma^2_i)^\iter}{\sigma_i}
\right)^2
}
+
2.5 \frac{\beta^2}{\alpha^4} (\epsilon_0 + \epsilon)\norm[2]{\vr_0}\\
&\mystackrel{(i)}{\leq}
\sqrt{
\sum_{i=1}^n
\left(
\innerprod{\vu_i}{\vy}
\frac{1 - (1 - \stepsize \sigma^2_i)^\iter}{\sigma_i}
\right)^2
}
+
\frac{1}{\alpha} \norm[2]{\vf(\mW_0, \vv_0)}
+
2.5 \frac{\beta^2}{\alpha^4} (\epsilon_0 + \epsilon)\norm[2]{\vr_0}\\
&\mystackrel{(ii)}{\leq}
\sqrt{
\sum_{i=1}^n
\left(
\innerprod{\vu_i}{\vy}
\frac{1 - (1 - \stepsize \sigma^2_i)^\iter}{\sigma_i}
\right)^2
}
+
\frac{\xi}{\alpha} \sqrt{n}.
\end{align*}
where (i) follows from $\vr_0 = \vf(\mW_0,\vv_0)-\vy$ and (ii) follows from
 $2.5 \frac{\beta^2}{\alpha^4} (\epsilon_0 + \epsilon)\norm[2]{\vr_0} \leq \frac{5}{16 \alpha} \norm[2]{\vr_0} \leq \frac{5}{16 \alpha} \left(\norm[2]{\vf(\mW_0,\vv_0)} + \norm[2]{\vy} \right)$, where we used $\epsilon_0 + \epsilon \leq 2 \epsilon = \frac{1}{8} \xi \frac{\alpha^3}{\beta^2}$, by definition of $\epsilon,\epsilon_0$, combined with 
 $\norm[2]{\vf(\mW_0,\vv_0)} \leq \xi/4 \sqrt{n}$.

\paragraph{Bound on change of coefficients:}
Finally, we establish the bounds on the change of the individual coefficients~\eqref{eq:littlewtr} and  \eqref{eq:littlevtr}. 
We start with the weights in the first layer, $\vw_r$. 
The gradient with respect to $\vw_r$ is given by $\nabla_{\vw_r} \mc L(\mW,\vv) = [\transp{\Jac}_1(\mW,\vv) ]_r \vr$, where $[\transp{\Jac}_1(\mW,\vv)]_r$ is the submatrix of the Jacobian multiplying with the weight $\vw_r$, and $\vr$ is the residual. 
Therefore, we obtain
\begin{align*}
\norm[2]{ \vw_{\iter,r} - \vw_{0,r} }
&=
\norm[2]{ \sum_{\iteralt=0}^{\iter-1} (\vw_{\iteralt+1,r} - \vw_{\iteralt,r}) } \\
&\leq
\sum_{\iteralt=0}^{\iter-1}
\eta
\norm[2]{ [\transp{\Jac}_1(\mW_\iteralt,\vv_\iteralt) ]_r \vr_{\iteralt} } \\
&\mystackrel{(i)}{\leq}
(\nu + \frac{4}{\alpha} \sqrt{n}) \frac{\sqrt{n}}{\sqrt{k}}
\eta
\sum_{\iteralt=0}^{\iter-1}
\norm[2]{ \vr_{\iteralt} } \\
&\mystackrel{(ii)}{\leq}
(\nu + \frac{4}{\alpha} \sqrt{n}) \frac{\sqrt{n}}{\sqrt{k}}
\eta
\sum_{\iteralt=0}^{\iter-1}
(1 - \eta \alpha^2)^\iteralt \norm[2]{ \vr_{0} } \\
&\mystackrel{(iii)}{\leq}
(\nu + \frac{4}{\alpha} \sqrt{n}) \frac{\sqrt{n}}{\sqrt{k}}
\frac{1}{\alpha^2} \norm[2]{ \vr_{0} }.
\end{align*}
Here, (i) follows from 
\begin{align}
\norm{[\transp{\Jac}_1(\mW_\iteralt,\vv_\iteralt) ]_r}
&=
\norm{ \frac{v_r}{\sqrt{k}} \diag(\relu'(\mX \vw_r)) \mX } \nonumber \\
&\leq
v_r \frac{\sqrt{n}}{\sqrt{k}} \nonumber \\
&\leq
(\nu + \frac{4}{\alpha} \sqrt{n}) \frac{\sqrt{n}}{\sqrt{k}},
\nonumber
\end{align}
where the last inequality follows from
$|v_{\iteralt,r} - v_{0,r}| \leq \norm[2]{ \vv_{\iteralt,r}  - \vv_{0,r} } \leq R \leq \frac{4}{\alpha} \sqrt{n}$, by \eqref{eq:boundradius}. 
Moreover, for (ii) we used that, by~\eqref{resnl}, the non-linear residuals converge geometrically, and (iii) follows from the formula for a geometric series. 
This conclude the proof of the bound~\eqref{eq:littlewtr}.

Analogously, we obtain
\begin{align*}
|v_r - v_{0,r}|
&\leq
\sum_{\iteralt=0}^{\iter-1}
\eta
| [\transp{\Jac}_2(\mW_\iteralt) ]_r \vr_{\iteralt} | \\
&\leq
\left(
O( \omega \sqrt{\log(n k / \delta)} )
+
\frac{4}{\alpha}
\sqrt{n}  \right)
\frac{\sqrt{n}}{\sqrt{k}}
\frac{1}{\alpha^2} \norm[2]{\vr_0}
\end{align*}
where we used that
\begin{align*}
\norm[2]{ \frac{1}{\sqrt{k}} \relu(\mX \vw_r) }
&\leq
\frac{1}{\sqrt{k}}
\left(
\norm[2]{\mX \vw_{0,r}}
+
\norm[2]{\mX (\vw_{0,r} - \vw_r)}
\right)
\\
&\leq
\frac{1}{\sqrt{k}} 
\left(
O( \omega \sqrt{n \log(n k / \delta)} )
+
\sqrt{n} \norm[2]{ \vw_{0,r} - \vw_r }
\right) \\
&\leq
\frac{\sqrt{n}}{\sqrt{k}} 
\left(
O( \omega \sqrt{\log(n k / \delta)} )
+
\sqrt{n} \frac{4}{\alpha^2}
\right).
\end{align*}
Here, the last inequality follows by using that the entries of $\mX \vw_{0,r}$ are not independent but $\mc N(0, \omega^2)$ distributed, and by taking an union bound over all entries of that vector and over all $\mX \vw_{0,r}$. 
This concludes the proof of the bound~\eqref{eq:littlevtr}.

\subsubsection{Proof of lemma~\ref{eq:lemchangeJac}}

First note that 
\begin{align}
\norm{\Jac(\mW,\vv) - \Jac(\mW',\vv')}
&\leq
\norm{\Jac_1(\mW,\vv) - \Jac_1(\mW',\vv')}
+
\norm{\Jac_2(\mW,\vv) - \Jac_2(\mW',\vv')} \nonumber \\
&\leq
\norm{\Jac_1(\mW,\vv) - \Jac_1(\mW,\vv') }
+
\norm{\Jac_1(\mW,\vv') - \Jac_1(\mW',\vv') } \nonumber \\
&+
\norm{\Jac_2(\mW,\vv) - \Jac_2(\mW',\vv')}.
\label{eq:overallboundjac}
\end{align}
In the reminder of the proof we bound the three terms above. 
We start by bounding the third term in~\eqref{eq:overallboundjac} as:
\begin{align*}
\norm{\Jac_2(\mW,\vv) - \Jac_2(\mW',\vv')}
&\leq
\frac{1}{\sqrt{k}}
\norm[F]{ \relu(\mX \mW) -  \relu(\mX \mW') } \\
&\leq
\frac{1}{\sqrt{k}}
\norm[F]{ \mX \mW - \mX \mW' } \\
&\leq
\frac{1}{\sqrt{k}}
\norm{\mX} \norm[F]{\mW - \mW' }.
\end{align*}
We proceed with bounding the first term in~\eqref{eq:overallboundjac} as:
\begin{align*}
\norm{\Jac_1(\mW,\vv) - \Jac_1(\mW,\vv') }^2
&= 
\norm{(\Jac_1(\mW,\vv) - \Jac_1(\mW,\vv')) \transp{(\Jac_1(\mW,\vv) - \Jac_1(\mW,\vv'))} } \\
&=
\frac{1}{k}
\norm{\relu'(\mX \mW) 
\diag((v_1-v_1')^2,\dots,(v_\iter - v_\iter')^2)
\transp{ \relu'(\mX \mW) } 
\odot
\mX \transp{\mX} }  \\
&\leq
\frac{1}{k}
\norm{\mX}^2
\max_j \norm[2]{ \relu'(\transp{\vx}_j \mW) \diag(\vv - \vv') }^2
 \\
&\leq
\frac{1}{k}
\norm[2]{\vv - \vv'}^2
\norm{\mX}^2.
\end{align*}
Next we establish below that with probability at least $1- n e^{- kq^2/2}$, the second term in in~\eqref{eq:overallboundjac} is bounded as
\begin{align}
\label{eq:finalboundjac1}
\norm{\Jac_1(\mW,\vv') - \Jac_1(\mW', \vv')}
&\leq
\frac{1}{\sqrt{k}}
\norm[\infty]{\vv'}
\norm{\mX} \sqrt{2q}
\end{align}
provided that 
\[
\norm{\mW - \mW'} 
\leq 
\sqrt{q} \frac{q}{2k} \omega
=
\omega 
\tilde R,
\]
where the last inequality follows from setting $q = (2 k\tilde R)^{2/3}$ (note that the assumption $\tilde R \leq \frac{1}{2} \sqrt{k}$ ensures $q \leq k$).
Putting those three bounds together in~\eqref{eq:overallboundjac} yields
\begin{align}
\norm{\Jac(\mW,\vv) - \Jac(\mW',\vv')}
&\leq
\norm{\mX}
\frac{1}{\sqrt{k}}
\left(
\norm[F]{\mW - \mW'}
+
\norm[2]{\vv - \vv'}
+
\norm[\infty]{\vv'} \sqrt{2} (2k\tilde R)^{1/3}
\right),
\end{align}
which established the claim.

It remains to prove~\eqref{eq:finalboundjac1}. Towards this goal, first note that
\begin{align}
\label{eq:overallboundjacsplit}
\norm{\Jac_1(\mW,\vv) - \Jac_1(\mW',\vv')}
\leq
\norm{\Jac_1(\mW,\vv) - \Jac_1(\mW,\vv') }
+
\norm{\Jac_1(\mW,\vv') - \Jac_1(\mW',\vv') }
\end{align}
We proceed with bounding the second term in the RHS of~\eqref{eq:overallboundjacsplit} as:
\begin{align}
\label{eq:reljacJaccact}
\norm{\Jac_1(\mW,\vv') - \Jac_1(\mW',\vv')}^2
&\leq
\frac{1}{k}
\norm[\infty]{\vv'}^2
\norm{\mX}^2
\max_j \norm{\sigma'(\transp{\vx}_j \mW)  -  \sigma'(\transp{\vx}_j \mW') }^2.
\end{align}
Because $\relu'$ is the step function, we have to bound the number of sign flips between the matrices $\mX \mW$ and $\mX \mW'$. For this we use the lemma below:

\begin{lemma}
Let $|\vv|_{\pi(q)}$ be the $q$-th smallest entry of $\vv$ in absolute value.
Suppose that, for all $i$, and $q \leq k$,
\[
\norm{\mW - \mW'} \leq \sqrt{q} \frac{|\transp{\vx}_i\mW'|_{\pi(q)}}{ \norm{\vx_i} }.
\]
Then
\[
\max_i \norm{ \sigma'( \transp{\vx}_i\mW ) - \sigma'( \transp{\vx}_i\mW' ) }
\leq
\sqrt{2q}.
\]
\end{lemma}

For the entries $\mW'$ being iid $\mc N(0,\omega^2)$, we note that, with probability at least $1 - n e^{- k q^2/2 }$, the $q$-th smallest entry of $\transp{\vx}_i \mW' \in \reals^k$ obeys
\begin{align}
\frac{|\transp{\vx}_i \mW'|_{\pi(q)} }{ \norm{\vx_i} } \geq \frac{q}{ 2 k} \omega
\quad \text{for all } i = 1,\ldots,n.
\end{align}
We are now ready to conclude the proof of the lemma. By equation~\eqref{eq:reljacJaccact},
\begin{align*}
\norm{\Jac_1(\mW,\vv') - \Jac_1(\mW', \vv')}
&\leq
\frac{1}{\sqrt{k}}
\norm[\infty]{\vv'}
\norm{\mX}
\max_j \norm{\sigma'(\transp{\vx}_j \mW)  -  \sigma'(\transp{\vx}_j \mW') } \\
&\leq
\frac{1}{\sqrt{k}}
\norm[\infty]{\vv}
\norm{\mX} \sqrt{2q} 
\end{align*}
provided that 
\[
\norm{\mW - \mW'} 
\leq 
\sqrt{q} \frac{q}{2k} \omega
\]
with probability at least $1- n e^{- kq^2/2}$.



%
%
%
%
%
%
%
%
%
%
%
%


\subsection{Proof of Lemma~\ref{lem:lemrademacher} (bound on the Rademacher complexity)}

Our proof follows that of a related result, specifically~\cite[Lem.~6.4]{arora_FineGrainedAnalysisOptimization_2019} which pertains to a two-layer ReLU network where only the first layer is trained and the second layers' coefficients are fixed.
Our goal is to bound the empirical Rademacher complexity
\begin{align*}
\setR_\setD(\setF_{Q,M})
=
\frac{1}{n}\EX{ \sup_{\mW \in \setW, \vv \in \setV}
\sum_{i=1}^n \epsilon_i
\frac{1}{\sqrt{k}}\sum_{r=1}^k v_r \relu(\transp{\vw}_r \vx_i) },
\end{align*}
where expectation is over the iid Rademacher random variables $\epsilon_i$, and where $\{\vx_1,\ldots,\vx_n\}$ are the training examples.

The derivation of the bound on the Rademacher complexity is based on the intuition that if the parameter $M$ of the constraint $\norm[2]{\vw_r - \vw_{0,r}} \leq \omega M$ is sufficiently small, then $\relu'(\transp{\vw}_r \vx_i)$ is constant for most $r$, because $|\transp{\vw}_{0,r} \vx_i|$ is bounded away from $\omega M$ with high probability, by anti-concentration of a Gaussian. 
For those coefficient vectors $r$ for which $\relu'(\transp{\vw}_r \vx_i)$ is constant, we have 
$\relu(\transp{\vw}_r \vx_i) = \relu'(\transp{\vw}_{0,r} \vx_i)\transp{\vw}_r \vx_i$. For the other coefficients, we can bound the difference of those two values as
\begin{align*}
\relu(\transp{\vw}_r \vx_i)
- \relu'(\transp{\vw}_{0,r} \vx_i)\transp{\vw}_r \vx_i
=&
\relu'(\transp{\vw}_r \vx_i) \transp{\vw}_r \vx_i
- \relu'(\transp{\vw}_{0,r} \vx_i ) \transp{\vw}_r \vx_i \\
=&
\relu'(\transp{\vw}_r \vx_i) \transp{\vw}_r \vx_i
-
\relu'(\transp{\vw}_{0,r} \vx_i) \transp{\vw}_{0,r} \vx_i \\
&+
\relu'(\transp{\vw}_{0,r} \vx_i) \transp{\vw}_{0,r} \vx_i
- 
\relu'(\transp{\vw}_{0,r} \vx_i) \transp{\vw}_r \vx_i \\
=&
\relu(\transp{\vw}_r \vx_i) 
-
\relu(\transp{\vw}_{0,r} \vx_i) +
\relu'(\transp{\vw}_{0,r} \vx_i) \innerprod{\vw_{0,r} - \vw_{r}}{\vx_i} \\
\leq&
2 \norm[2]{\vw_r - \vw_{0,r}} \norm[2]{\vx_i} \\
\leq&
2 \omega M,
\end{align*}
where the last inequality holds for $\mW \in \setW$. 
It follows that
\begin{align}
\setR_\setD(\setF_{Q,M})
%
%
&\leq
\frac{1}{n}\EX{ \sup_{\mW \in \setW, \vv \in \setV} 
\sum_{i=1}^n \epsilon_i \frac{1}{\sqrt{k}} \sum_{r=1}^k v_r 
\relu'(\transp{\vw}_{r,0} \vx_i)
\transp{\vw}_r \vx_i} \nonumber \\
&\hspace{0.5cm}+ 
\frac{2 M}{n \sqrt{k} } \sum_{i=1}^n \sum_{r=1}^k  v_r \ind{ \relu'(\transp{\vw}_{r,0} \vx_i ) \neq  \relu'(\transp{\vw}_{r} \vx_i) }
\nonumber \\
&=
\frac{1}{n} \EX{ \sup_{\mW \in \setW, \vv \in \setV}
\transp{\vepsilon}
\Jac_1(\mW_0,\vv) \vw
}
+ 
\frac{2 \omega M}{n \sqrt{k} } 
\sum_{i=1}^n \sum_{r=1}^k v_r \ind{ \relu'(\transp{\vw}_{r,0} \vx_i ) \neq  \relu'(\transp{\vw}_{r} \vx_i) },
\label{eq:boundadf}
\end{align}
where $\Jac_1$ is the Jacobian defined in~\eqref{eq:jacobian}, and where we use $\vw = \vectorize{\mW} \in \reals^{d k}$ for the vectorized version of the matrix $\mW$ with a slight abuse of notation. 
With this notation, we can bound the first term in~\eqref{eq:boundadf} by
\begin{align}
\frac{1}{n} \EX{ \sup_{\mW \in \setW, \vv \in \setV}
\transp{\vepsilon}
\Jac_1(\mW_0,\vv) \vw
}
\hspace{-4cm}&\hspace{+4cm}\mystackrel{(i)}{=}
\frac{1}{n} \EX{ \sup_{\mW \in \setW, \vv \in \setV}
\transp{\vepsilon}
\left(
\Jac_1(\mW_0,\vv) \vw
-
\Jac_1(\mW_0,\vv_0) \vw_0
\right)
}
\nonumber \\
&=
\frac{1}{n} \EX{ \sup_{\mW \in \setW, \vv \in \setV}
\transp{\vepsilon}
\left(
\Jac_1(\mW_0,\vv) \vw
-\Jac_1(\mW_0,\vv) \vw_0
+\Jac_1(\mW_0,\vv) \vw_0
-\Jac_1(\mW_0,\vv_0) \vw_0
\right)
}
\nonumber \\
&=
\frac{1}{n} \EX{ \sup_{\mW \in \setW, \vv \in \setV}
\transp{\vepsilon}
\left(
\Jac_1(\mW_0,\vv) (\vw - \vw_0)
+
\Jac_2(\mW_0) (\vv - \vv_0)
\right)
}
\nonumber \\
&=
\frac{1}{n} \EX{ \sup_{\mW \in \setW, \vv \in \setV}
\transp{\vepsilon}
\left(
\Jac_1(\mW_0,\vv_0) (\vw - \vw_0)
+
\Jac_1(\mW_0,\vv - \vv_0) (\vw - \vw_0)
+
\Jac_2(\mW_0) (\vv - \vv_0)
\right)
}
\nonumber \\
&\mystackrel{(ii)}{\leq}
\frac{1}{n} \EX{
\norm[2]{\transp{\vepsilon}
\Jac_1(\mW_0,\vv_0)} } Q
+
\frac{1}{n} \EX{
\norm[2]{\transp{\vepsilon}
\Jac_2(\mW_0)} 
} Q 
+
\sqrt{k} \nu \omega M^2
\nonumber \\
&\mystackrel{(iii)}{\leq}
\frac{1}{n}Q(\nu + \omega) \sqrt{n} + \sqrt{k} \nu \omega M^2.
\label{eq:secoundonradc}
\end{align}
Here, 
equality (i) follows because $\vepsilon \Jac_1(\mW_0,\vv_0)\vw_0$ has zero mean, 
inequality (ii) follows from the Cauchy-Schwarz inequality as well as from
\[
\norm[2]{\Jac_1(\mW_0,\vv - \vv_0) (\vw-\vw_0)} \leq \frac{1}{\sqrt{k}} \norm{\mX} \sum_{r=1}^k |v_r - v_{0,r}|\norm[2]{\vw_r - \vw_{0,r}} \leq \sqrt{k} \nu \omega M^2 \sqrt{n},
\]
and inequality (iii) follows from $\EX{\norm[2]{\mA \vepsilon}} \leq \sqrt{ \EX{ \norm[2]{\mA \vepsilon}^2 } } = \norm[F]{\mA}$, by Jensen's inequality, and from the bounds $\norm[F]{\Jac_1(\mW_0,\vv_0)} \leq \nu$ and $\norm[F]{\Jac_2(\mW_0)} \leq \omega$, which holds with probability at least $1-\delta$ provided that $\sqrt{\log(2n/\delta) / 2 k} \leq 1/2$, which in turn holds by assumption. 

We next upper bound the second term in~\eqref{eq:boundadf}. 
Following the argument in~\cite[Proof of Lemma 5.4]{arora_FineGrainedAnalysisOptimization_2019}, we get 
\begin{align}
\sum_{i=1}^n \sum_{r=1}^k v_r \ind{ \relu'(\transp{\vw}_{r,0} \vx_i ) \neq  \relu'(\transp{\vw}_{r} \vx_i) }
\leq
2\nu
kn \left( M + \sqrt{\frac{\log(2/\delta)}{2k }} \right),
\label{eq:secoundonradsecond}
\end{align}
with probability at least $1-\delta$. Here, we used that $v_r \leq |v_{0,r}| + |v_r - v_{0,r}| \leq 2\nu$. 
Putting the bounds on the first and second term in~\eqref{eq:boundadf} (given by inequality~\eqref{eq:secoundonradc} and inequality~\eqref{eq:secoundonradsecond}) together, we get that, with probability at least $1-\delta$, the Rademacher complexity is upper bounded by 
\begin{align*}
\setR_\setD(\setF)
\leq
\frac{Q}{\sqrt{n}} (\nu+\omega)
+
\nu \omega
\left( 5M^2 \sqrt{k}  + 4M \sqrt{\log(2/\delta)} \right)
\end{align*}
which concludes our proof.

\end{document}